\documentclass[10pt,twocolumn,letterpaper]{article}

\usepackage{cvpr}              

\usepackage{kotex}
\usepackage{lipsum}
\usepackage{pifont}
\usepackage[normalem]{ulem}

\usepackage{mathtools}

\usepackage{array}
\newcolumntype{P}[1]{>{\centering\arraybackslash}p{#1}}
\newcolumntype{M}[1]{>{\centering\arraybackslash}m{#1}}
\newcolumntype{L}[1]{>{\hspace{0.5em}\raggedright\arraybackslash}m{#1}}
\newcolumntype{R}[1]{>{\raggedleft\arraybackslash}m{#1}}

\usepackage{xspace}

\makeatletter
\DeclareRobustCommand\onedot{\futurelet\@let@token\@onedot}
\def\@onedot{\ifx\@let@token.\else.\null\fi\xspace}

\def\eg{\emph{e.g}\onedot} 
\def\ie{\emph{i.e}\onedot}

\makeatother

\usepackage{caption}
\usepackage{tabularx,booktabs}
\newcolumntype{Y}{>{\centering\arraybackslash}X}
\usepackage{colortbl}
\usepackage{multirow}
\usepackage{soul}
\usepackage{wrapfig}
\let\drule\relax
\newcommand{\drule}{\specialrule{0.2pt}{1pt}{1pt}%
            \specialrule{0.2pt}{0pt}{\belowrulesep}%
            }

\usepackage{algorithm,algpseudocode}
\usepackage[algo2e]{algorithm2e} 
\definecolor{commentcolor}{RGB}{110,154,155}
\definecolor{defcolor}{RGB}{225,81,145}
\let\PyComment\relax
\let\PyCode\relax
\let\PyDef\relax
\newcommand{\PyComment}[1]{\footnotesize \ttfamily \textcolor{commentcolor}{\# #1}}  
\newcommand{\PyCode}[1]{\footnotesize \ttfamily \textcolor{black}{#1}} 
\newcommand{\PyDef}[1]{\footnotesize \ttfamily \textcolor{defcolor}{#1}} 

\usepackage{titletoc}

\let\origparagraph\paragraph
\renewcommand{\paragraph}[1]{\noindent\textbf{#1 \hspace{0.1cm}}}

\usepackage{amsmath,amsfonts,bm}

\def\eqref#1{equation~\ref{#1}}

\def\1{\bm{1}}

\DeclareMathAlphabet{\mathsfit}{\encodingdefault}{\sfdefault}{m}{sl}
\SetMathAlphabet{\mathsfit}{bold}{\encodingdefault}{\sfdefault}{bx}{n}

\def\gA{{\mathcal{A}}}

\def\gF{{\mathcal{F}}}

\newcommand{\E}{\mathbb{E}}

\newcommand{\R}{\mathbb{R}}

\newcommand\norm[1]{\left\lVert#1\right\rVert}

\definecolor{cvprblue}{rgb}{0.21,0.49,0.74}
\usepackage[pagebackref,breaklinks,colorlinks,allcolors=cvprblue]{hyperref}

\usepackage[capitalize]{cleveref}
\Crefformat{figure}{Fig.~#2#1#3}
\Crefformat{equation}{Eq.~#2#1#3}
\Crefformat{table}{Tab.~#2#1#3}

\def\paperID{****} 
\def\confName{CVPR}
\def\confYear{2026}

\title{Concept-Aware LoRA for Domain-Aligned Segmentation Dataset Generation}

\newcommand*{\affmark}[1][*]{\textsuperscript{#1}}
\newcommand\blfootnote[1]{%
  \begingroup
  \renewcommand\thefootnote{}\footnote{#1}%
  \addtocounter{footnote}{-1}%
  \endgroup
}

\author{
Minho Park\affmark[1]$^\text{*}$ \; Sunghyun Park\affmark[2] \; Jungsoo Lee\affmark[2] \; Hyojin Park\affmark[2] \; \\ 
Kyuwoong Hwang\affmark[2] \; Fatih Porikli\affmark[2] \; Jaegul Choo\affmark[1] \; Sungha Choi\affmark[3]$^\dagger$ \; \vspace{0.12cm}\\
\affmark[1]KAIST \; \affmark[2]Qualcomm AI Research$^\ddagger$ \; \affmark[3]Kyung Hee University\\
\texttt{\footnotesize\{m.park, jchoo\}@kaist.ac.kr} \ \texttt{\footnotesize sunghac@khu.ac.kr} \\[-1pt] \texttt{\footnotesize\{sunpar, jungsool, hyojinp, kyuwoong, fporikli\}@qti.qualcomm.com} \;
\vspace{-0.1cm}
}

\DeclareRobustCommand{\hlblue}[1]{{\sethlcolor{blue!20}\hl{#1}}}
\DeclareRobustCommand{\hlred}[1]{{\sethlcolor{red!20}\hl{#1}}}

\definecolor{darkgreen}{RGB}{25,200,25}

\definecolor{revcolor}{RGB}{245,135,0}

\begin{document}
\maketitle

\blfootnote{\hspace{-0.4cm}$^*$Work done during an internship at Qualcomm AI Research.}
\blfootnote{\hspace{-0.4cm}$^\dagger$Corresponding author. Work done while at Qualcomm AI Research.} \blfootnote{\hspace{-0.4cm}$^\ddagger$Qualcomm AI Research is an initiative of Qualcomm Technologies, Inc.}

\begin{abstract}
This paper addresses the challenge of data scarcity in semantic segmentation by generating datasets through text-to-image (T2I) generation models, reducing image acquisition and labeling costs.
Segmentation dataset generation faces two key challenges: 1) aligning generated samples with the target domain and 2) producing informative samples beyond the training data.
Fine-tuning T2I models can help generate samples aligned with the target domain.
However, it often overfits and memorizes training data, limiting their ability to generate diverse and well-aligned samples.
To overcome these issues, we propose Concept-Aware LoRA (CA-LoRA), a novel fine-tuning approach that selectively identifies and updates only the weights associated with necessary concepts (\eg, style or viewpoint) for domain alignment while preserving the pretrained knowledge of the T2I model to produce informative samples.
We demonstrate its effectiveness in generating datasets for urban-scene segmentation, outperforming baseline and state-of-the-art methods in in-domain (few-shot and fully-supervised) settings, as well as in domain generalization tasks, especially under challenging conditions such as adverse weather and varying illumination, further highlighting its superiority.

\end{abstract}

\vspace{-10pt}
\section{Introduction}
The amount of labeled data is crucial for achieving high performance in semantic segmentation.
However, collecting diverse image samples, especially in rare or complex scenarios, and providing pixel-level annotations is both labor-intensive and time-consuming.
To address this data scarcity, recent studies~\cite{zhang2021datasetgan, baranchuk22label, datasetdm, wu2023diffumask} have leveraged text-to-image (T2I) generation models~\cite{ldm,imagen,sdxl} to synthesize training data for semantic segmentation.
These studies utilize the rich generative capabilities of T2I models to produce promising image-label pairs.
In addition, the text-driven controllability of T2I models enables targeted generation of underrepresented distributions.
This has proven particularly effective for augmenting labeled datasets, alleviating data scarcity, and generating specific distributions such as balancing class~\cite{shin2023fill} or weather conditions~\cite{jia2023dginstyle}.

\begin{figure*}[t]
\centering
\includegraphics[width=0.98\textwidth]{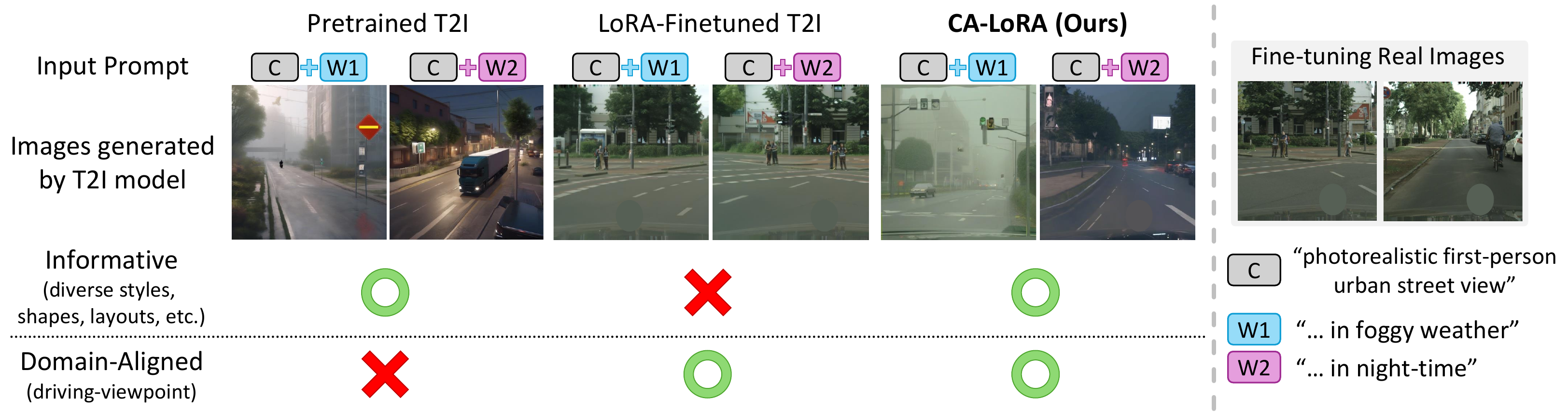}
\vspace{-0.25cm}
\caption{
\textbf{Motivation of Concept-Aware LoRA (CA-LoRA).}
Pretrained T2I models generate informative images but struggle with viewpoint alignment.
LoRA fine-tuning on Cityscapes enables driving-viewpoint generation but leads to overfitting to the Cityscapes style and content.
We aim to learn only the desired concept (\eg, viewpoint) for generating domain-aligned, informative samples.
}
\label{fig:1_intro_motivation}
\vspace{-0.5cm}
\end{figure*}

There are two primary challenges in generating segmentation datasets using image generation models: (1) \textit{generating informative datasets beyond the original training data}, and (2) \textit{achieving alignment with the target domain}, as illustrated in \cref{fig:1_intro_motivation}.
However, prior approaches have often overlooked key aspects of these challenges.
Early works on segmentation dataset generation that do not utilize pretrained T2I models~\cite{zhang2021datasetgan, li2022bigdatasetgan, baranchuk22label, park2023learning} typically produce domain-aligned samples, as the generative models are trained solely on the target dataset.
Yet, due to the lack of external knowledge, they are limited in their ability to generate informative samples beyond the training distribution.

In contrast, recent studies~\cite{datasetdm, wu2023diffumask, yang2024freemask, jia2023dginstyle} leverage pretrained T2I models such as Stable Diffusion, which are trained on large-scale datasets like LAION-5B~\cite{laion5b}, enabling them to incorporate external information.
However, these methods often use the pretrained models without task-specific fine-tuning for segmentation, which leads to poor alignment with the target domain.

To address this, we apply existing fine-tuning methods~\cite{hu2022lora} to align the image domain of pretrained generative models with the target domain.
However, as illustrated in \Cref{fig:1_intro_motivation}, even LoRA-based fine-tuning tends to overfit and memorize the training data, limiting the model's ability to generate diverse and informative samples.
This overfitting arises because the model learns all concepts present in the training data, such as viewpoint, style, object shape, and layout, regardless of whether they are necessary for domain alignment.
To overcome this limitation, we argue that fine-tuning should selectively focus on only the essential concepts (e.g., viewpoint or style) required for domain alignment, while preserving the pretrained T2I model’s capacity to generate informative samples.
Moreover, since the relevant concepts for alignment may vary across different problem settings, we propose a flexible fine-tuning method that adapts to the necessary concepts for \textit{generating urban-scene segmentation datasets}.
In in-domain segmentation settings, where the source and target domains are aligned (\eg, clear-day conditions), the T2I model primarily needs to learn the \textit{clear-day style} of the training data.
In contrast, in domain generalization (DG) settings, where the target domain is unseen (\eg, rainy-day conditions), focusing on the \textit{driving viewpoint} is often more effective, as style alignment is no longer feasible.
Our method supports both settings by selectively tuning the relevant concept, enabling the generation of datasets that are both informative and well aligned with the task requirements.

We propose Concept-Aware LoRA (CA-LoRA), a fine-tuning method that \textit{automatically identifies and updates only the weights} of T2I models associated with the desired concepts, while keeping the remaining weights frozen to preserve the pretrained knowledge.
This allows the model to \textit{learn only the specified concepts}, enabling it to generate images that are both well-aligned with the target domain and more diverse and informative.
For example, when the desired concept is the \textit{viewpoint}, our method fine-tunes the model to learn only the viewpoint from the given dataset, allowing it to generate images with varying styles, object shapes, and layouts---capabilities that standard LoRA fine-tuned models lack, as they tend to overfit to the training style and fail to generalize beyond it, even with varied text prompts.
Consequently, our fine-tuned T2I models can generate specific styles based on user input, making them particularly effective in DG settings~\cite{choi2021robustnet,hoyer2022hrda,hoyer2022daformer}.

Our analysis shows that the proposed CA-LoRA can exclusively learn the specified concepts, and we demonstrate the effectiveness of the generated datasets for urban-scene segmentation by comparing them with state-of-the-art dataset generation methods~\cite{datasetdm,jia2023dginstyle,brooks2023instructpix2pix,benigmim2023one} in both in-domain and domain generalization settings.

Our contributions are threefold:
\begin{enumerate}
    \item We propose Concept-Aware LoRA, a novel fine-tuning method that selectively identifies and updates only the weights related to the necessary concepts, reducing overfitting and preserving pretrained knowledge.
    \item Concept-Aware LoRA enables T2I models to generate well-aligned and informative samples beyond training data. This addresses data scarcity by generating image-label pairs from underrepresented distributions (\eg, adverse weather), improving segmentation performance.
    \item Our method achieves state-of-the-art performance across various settings, including a +2.30\% mIoU improvement in few-shot and +1.34\% in fully supervised settings on Cityscapes, as well as an average gain of +1.53\% mIoU on domain generalization benchmarks.
\end{enumerate}
%

\begin{figure*}[t]
\centering
\includegraphics[width=0.95\textwidth]{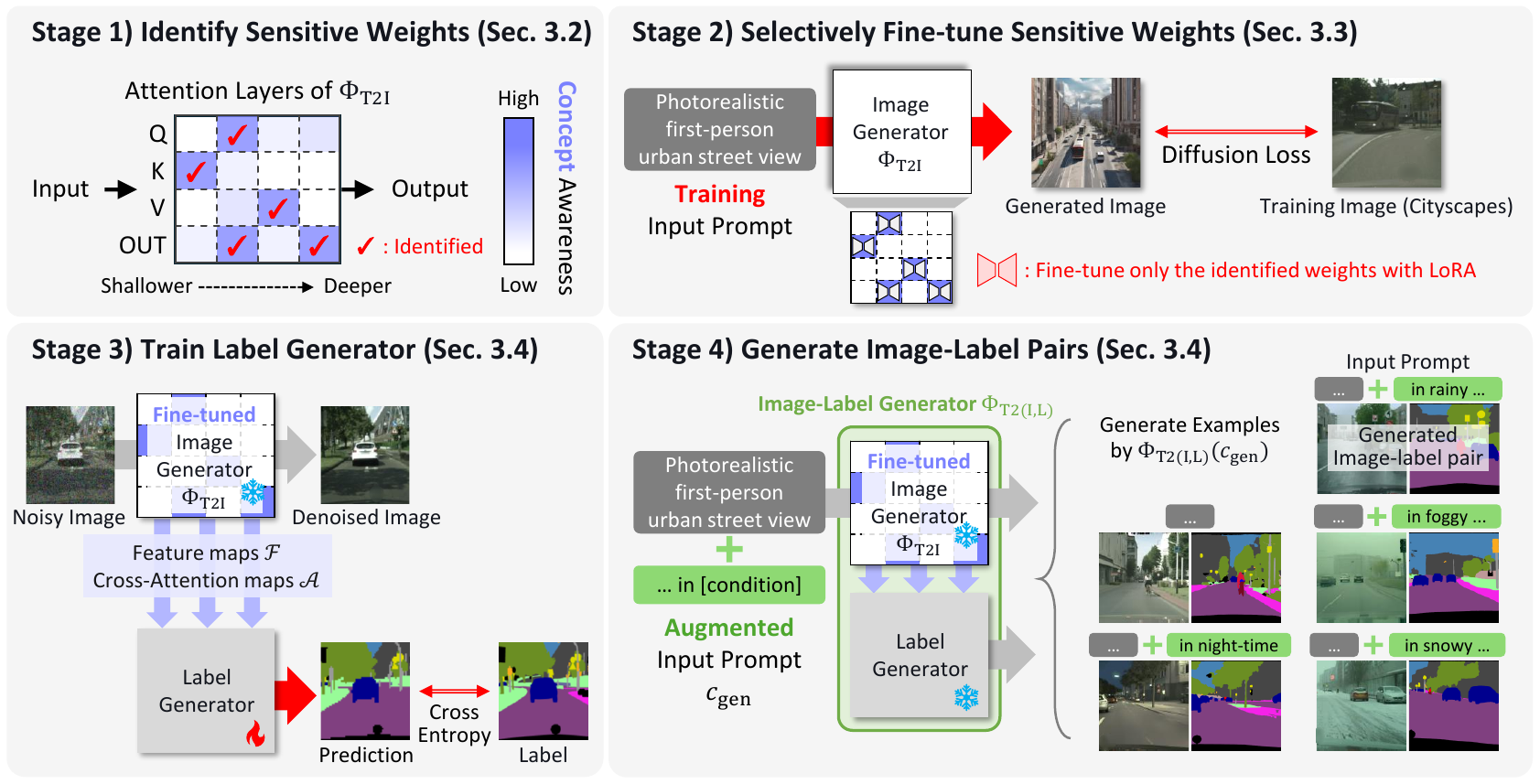}
\vspace{-0.25cm}
\caption{
\textbf{Overview of our framework for generating an urban-scene segmentation dataset by learning the Cityscapes viewpoint.}
The process consists of four stages: (1)~identifying sensitive weights for a specific concept, (2)~selectively fine-tuning them with LoRA, (3)~training a label generator using features from T2I model, and (4)~generating diverse image-label pairs with augmented prompts.
}
\label{fig:1_intro_overview}
\vspace{-0.35cm}
\end{figure*}

\section{Related Work}
\subsection{Efficient Fine-Tuning of T2I Models}
\vspace{-3pt}

Recent advancements in diffusion architectures~\cite{ddpm, ldm} and large-scale image-text dataset~\cite{laion5b} have enabled high-quality T2I models~\cite{imagen, dalle2, sdxl, sd3}.
The quality of images generated by these models has led researchers to personalize them to produce specific concepts or styles~\cite{ruiz2023dreambooth, gal2022image}. 
To achieve better customization, parameter-efficient fine-tuning (PEFT) methods~\cite{hu2022lora, liu2024dora, hayou2024lora+, kopiczko2023vera, ding2023sparse, he2022sparseadapter} have been proposed.
While existing PEFT methods aim to prevent overfitting and enable efficient training, they struggle to disentangle irrelevant concepts during fine-tuning, as they may still equally affect all layers.
Thus, several studies~\cite{guo2019spottune, choi2022improving, lee2022surgical} have shown that fine-tuning manually selected layers outperforms full fine-tuning, especially with smaller datasets.
Additionally, recent work on Stable Diffusion~\cite{wang2024instantstyle, xing2024csgo, basu2024mechanistic} identifies control blocks for specific visual attributes by ablating each block manually.
In contrast, our approach aims to automate this process, allowing precise, fine-grained updates to only the most crucial weights.

\subsection{Segmentation Dataset Generation}
\vspace{-3pt}

Generating semantic segmentation datasets has been deeply explored, as pixel-wise annotations are particularly expensive.
One classical approach to generating segmentation pairs is to create corresponding label maps for the images produced by image generation models~\cite{zhang2021datasetgan, li2022bigdatasetgan, baranchuk22label, park2023learning}.
Specifically, they utilized the intermediate features of the generative models (\eg, GAN, DM) during image generation, which contains rich semantical information of the generated image.
By leveraging these rich generative features, a label generator can be trained with minimal labeled data.

Recent studies have focused on leveraging the pretrained knowledge of T2I models~\cite{datasetdm, wu2023diffumask, nguyen2024dataset_diffusion, benigmim2023one, gong2023prompting, kaplan2024domain}, but often overlook the alignment of the generated images with the target domain (\eg, style, viewpoint).
This paper explores the impact of fine-tuning T2I models for segmentation dataset generation, focusing on ensuring better image alignment.

\begin{figure*}[t]
\centering
\includegraphics[width=0.95\textwidth]{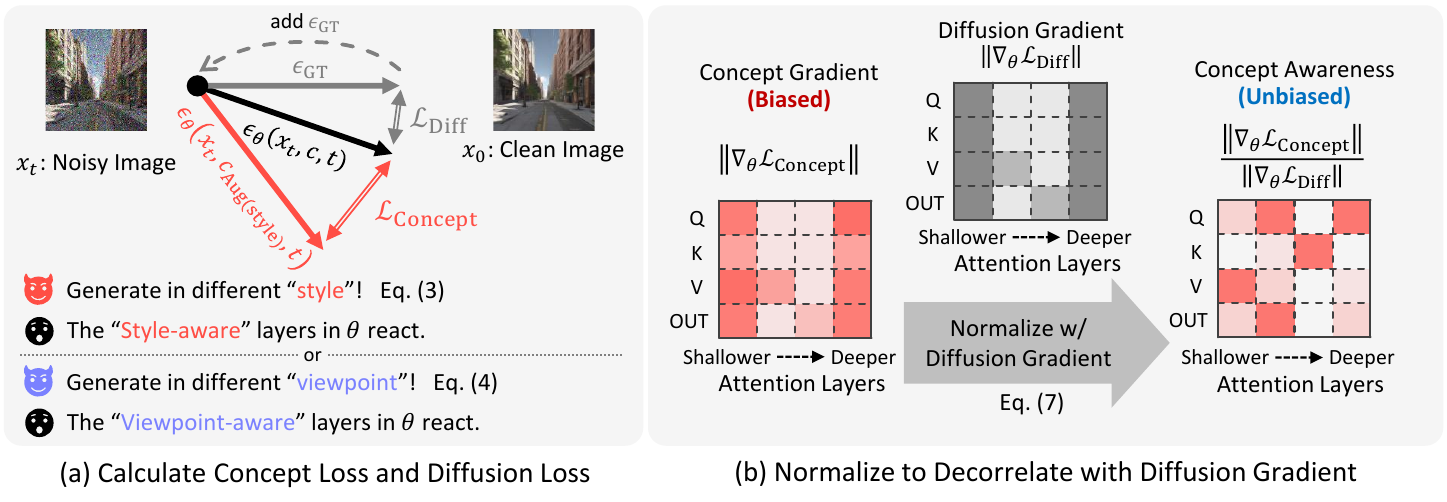}
\vspace{-0.15cm}
\caption{
\textbf{Overview of measuring concept awareness.}
(a) We design the concept loss ($\mathcal{L}_\text{Concept}$) with the concept-augmented captions ($c_\text{Aug}$), and the original diffusion loss ($\mathcal{L}_\text{Diff}$) with the added noise $\epsilon$.
The concept-augmented captions can be changed according to the desired concept (\eg, style, viewpoint).
(b) While each concept gradient represents the reaction of the concept, it has to be normalized with the original diffusion gradient to assess the increased ratio of each layer.
}
\label{fig:3_method_1_measure}
\vspace{-0.45cm}
\end{figure*}

\section{Method}
\subsection{Overall Framework}

\Cref{fig:1_intro_overview} shows our dataset generation framework.
First, we identify weights sensitive to the desired concept to be learned from the source dataset (Stage 1).
Next, Concept-Aware LoRA selectively fine-tunes the top-$k$\% sensitive weights of the T2I model (Stage 2).
Then, a label generator is trained on a labeled dataset using rich generative features from the T2I model (Stage 3).
Finally, diverse image-label pairs are generated with augmented prompts according to the problem settings (Stage 4).

\subsection{Identify Sensitive Weights to Desired Concept}
\label{sec:3_method_measure_score}

To identify the sensitive weights for a specific concept (\eg, style, viewpoint), we design an objective function, Concept loss ($\mathcal{L}_\text{Concept}$), which can be flexibly changed according to the desired concept.
As shown in \Cref{fig:3_method_1_measure} (a), Concept loss is applied to the noisy image to enforce modifying the concept.
For example, when the Concept loss forces the T2I model to modify \textit{style} (\eg, photorealistic $\rightarrow$ sketch),
its gradient can be used to identify \textit{style-aware weights}.

First, for the Concept loss input, we generate a few images $x_0$ using the T2I model $\Phi_\text{T2I}$ parameterized by $\theta$ and the text prompt $c$ (\ie, $x_0 = \Phi_\text{T2I} (c; \theta)$).
Random Gaussian noise $\epsilon$ is then added to the images based on the pre-defined timestep $t$ and the timestep scheduling coefficient $\bar{\alpha}_t$.
\begin{equation}
    x_t =\sqrt{\bar{\alpha}_t} x_0 + \sqrt{1 - \bar{\alpha}_t} \epsilon,
    \quad
    \epsilon \sim \mathcal{N}(0, \mathbf{I})
\end{equation}

Then, we prepare the augmented versions of the original text prompt according to style and viewpoint.
For example, when the original text prompt is given as

\vspace{-18pt}
\begin{equation}
c = \text{``Photorealistic first-person urban street view''},
\end{equation}

\vspace{-6pt}
\noindent we design \hlred{style-} and \hlblue{viewpoint-}augmented prompts as follows:

\vspace{-12pt}
{
\footnotesize
\begin{align}
c_\text{Aug\hlred{(Style)}} &= \text{``\hlred{Sketch of} first-person urban street view''}, \\[-2pt]
c_\text{Aug\hlblue{(Viewpoint)}} &= \text{``Photorealistic urban street \hlblue{in top-down view}''}.
\end{align}
}

We then use denoising prediction generated using the augmented caption as pseudo-ground truth to guide image modification. 
Concept loss ($\mathcal{L}_\text{Concept}$) is defined as follows, similar to the original diffusion loss.

\vspace{-13pt}
\begin{equation}
    \mathcal{L}_\text{Concept} := \norm{\epsilon_\theta (x_t, c, t) - \texttt{sg} [ \epsilon_\theta (x_t, c_\text{Aug}, t) ]}^2_2,
\end{equation}

\vspace{-2pt}
\noindent where $\texttt{sg}$ denotes the stop-gradient operation.
Initially, we quantify concept awareness using the RMS norm of the concept loss gradient ($\norm{\nabla_\theta \mathcal{L}_\text{Concept}}$) to compare gradients across layers with different dimensions.
However, the gradients show a significant positional bias across layers (see \Cref{appn:analysis_concept_awareness} for details).
To address this, we scale the gradient magnitude for each layer by computing the ratio between the gradients of the concept loss and the original diffusion loss ($\norm{\nabla_\theta \mathcal{L}_\text{Diff}}$), referred to as \emph{concept awareness}.

\vspace{-5pt}
\begin{equation}
    \mathcal{L}_\text{Diff} := \norm{\epsilon_\theta (x_t, c, t) - \epsilon }^2_2
\end{equation}
\vspace{-9pt}
{
\small
\begin{equation}
    \label{eq:3_method_awareness}
    \text{Concept-Awareness} (\theta) := \E_{x_0, \epsilon, c_\text{Aug}} \left[ \frac{\norm{\nabla_\theta \mathcal{L}_\text{Concept}} }{\norm{\nabla_\theta \mathcal{L}_\text{Diff}}
    }
    \right]
\end{equation}
}
We average the ratio over images ($x_0$), noise ($\epsilon$), and augmented prompts ($c_\text{Aug}$) to ensure statistical robustness.

Using the measured concept awareness values computed for each weight, we group and average them by attention projection layers (Q, K, V, and OUT) to quantify the concept awareness of each projection layer in the T2I model.



\subsection{Concept-Aware LoRA (CA-LoRA)}
\label{sec:3_method_slora}

After identifying concept-sensitive weights, we then apply LoRA updates only to these selected layers.
\emph{Concept-Aware LoRA (CA-LoRA)}, our parameter-efficient fine-tuning method, selectively updates only the concept-relevant weights while keeping the remaining parameters frozen.
This selective adaptation allows the model to learn the desired concept, yet still retain the pretrained T2I model’s knowledge across other, unrelated concepts.

\begin{figure}[t]
    \centering
    \includegraphics[width=\linewidth]{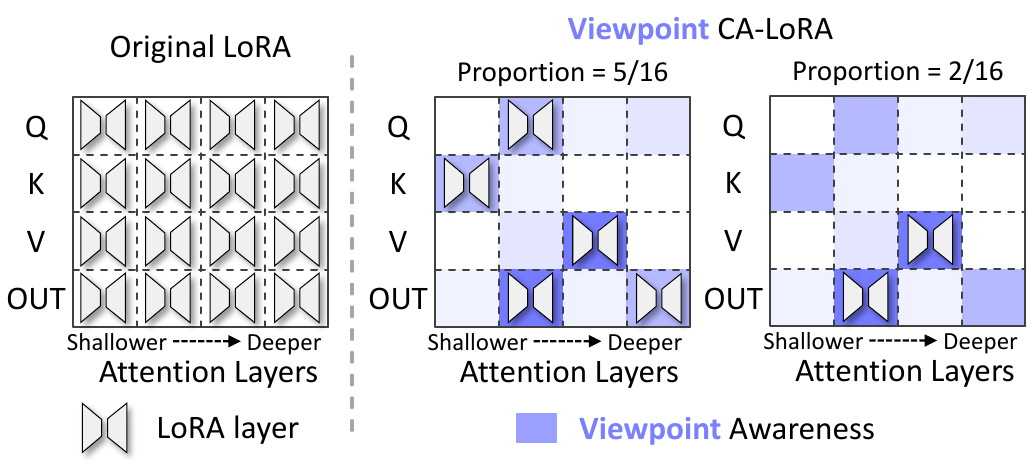}
    \vspace{-15pt}
    \caption{
    Illustration of CA-LoRA. Unlike the original LoRA, our CA-LoRA selectively attaches LoRA layers in a specified proportion to projection layers sensitive to the desired concept.
    }
    \label{fig:3_method_2_slora}
    \vspace{-10pt}
\end{figure}

We begin with Low-Rank Adaptation (LoRA)~\cite{hu2022lora}, a widely used fine-tuning method for large-scale pretrained T2I models.
For a pretrained weight matrix $W_0 \in \R^{d\times k}$, LoRA constrains updates using low-rank decomposition, $W_0 + \Delta W = W_0 + BA$, where $B \in \R^{d\times r}, A\in \R^{r\times k}$ and the rank $r \ll \text{min}(d, k)$.
In the literature on fine-tuning T2I models, LoRA is commonly applied to all projection layers (Q, K, V, OUT) in each attention block of the pretrained T2I model~\cite{ldm, hu2022lora} (left in \Cref{fig:3_method_2_slora}).
LoRA and its extensions~\cite{zhang2023adalora,lora_sp,gs_lora,tied_lora} aim to improve parameter efficiency for better adaptation to downstream tasks (\eg, by refining low-rank matrix decomposition).
However, selective learning approaches for a specific concept from given datasets have not been extensively explored.

CA-LoRA selectively fine-tunes the top-$k$\% of projection layers based on concept awareness (\cref{eq:3_method_awareness}), enabling it to learn only the desired concept.
Our experiments examine CA-LoRA in terms of viewpoint and style awareness.
The selected proportions are user-adjustable, allowing control over the extent of fine-tuning, as shown in \Cref{fig:3_method_2_slora} (right).
In the following sections, we refer to \hlred{Style CA-LoRA} and \hlblue{Viewpoint CA-LoRA} as CA-LoRA fine-tuning based on style and viewpoint awareness, respectively.

\subsection{Label Generator and Dataset Generation}
\label{sec:3_method_generation}

\paragraph{Training Label Generator}
We train an additional lightweight label generator following DatasetDM~\cite{datasetdm} to produce a segmentation label corresponding to the image.
First, given the real image $x$ from the training set, we add noise to the real image $x$ to synthesize noisy image $x_t$.
Then, we extract multi-scale generative features (feature maps $\gF$, cross-attention maps $\gA$) from the T2I generator $\epsilon_\theta (x_t, c, t)$, which contain lots of semantic information~\cite{zhang2021datasetgan,wu2023diffumask,khani2023slime}.
Last, the Mask2Former-shaped~\cite{mask2former} label generator receives the generative features $\gF, \gA$ and predicts the label maps as illustrated in Stage 3 of \Cref{fig:1_intro_overview}.
In summary, the entire process can be encapsulated by the text-to-(image-and-label) generator, denoted as $\Phi_\text{T2(I,L)}$.

\paragraph{Reducing Domain Gap in Generative Feature}
Importantly, a notable domain gap exists between the training images and the generated images from the pretrained T2I models, as shown in \Cref{fig:1_intro_motivation}.
This discrepancy leads to a domain gap in the generative features at inference, whose statistics differ substantially from those at training, thereby \textit{degrading the performance of the label generator.}
Unlike DatasetDM~\cite{datasetdm}, we train the label generator using the \textit{fine-tuned T2I model with CA-LoRA.}
Fine-tuning the T2I model on the training images greatly reduces this domain gap, allowing the label generator to receive generative features at inference that are statistically consistent with those seen during training, which in turn results in a substantial improvement in image-label alignment.

\paragraph{Generating Diverse Image-Label Paired Dataset}
Generating adverse weather or light conditions (\eg, foggy, night-time) is crucial for improving domain generalization in urban-scene segmentation, as described in DGInStyle~\cite{jia2023dginstyle}.
To achieve this, we append weather conditions as illustrated in Stage 4 of \Cref{fig:1_intro_overview}.
However, fine-tuning T2I models with the original LoRA often leads to overfitting to undesired concepts from the source dataset (\eg, clear-day style) and compromises controllability over textual augmentation.
In contrast, our CA-LoRA fine-tunes only the viewpoint-aware layers, enabling selective learning of viewpoint concepts while preserving the T2I model's text adherence in style augmentation.
This capability plays a critical role in generating diverse image-label pairs.
Also, we vary class names in the prompt (\eg, ``... with car, person, etc.'') to provide additional diversity.
In summary, the generation prompts are formulated as $c_\text{Gen} =$ \textit{``Photorealistic first-person urban street view with [class names] in [weather].''}.

\begin{table*}[t!]
\centering
\caption{
\textbf{In-domain segmentation performance across various fractions of the Cityscapes dataset (mIoU).}
The first row presents the baseline model trained solely on the real dataset.
We visualize the performance improvement relative to the baseline alongside each score.
}
\label{tab:exp_cityscapes}
\vspace{-5pt}
\resizebox{0.78\textwidth}{!}{
\begin{tabular}{@{\hspace{0.1cm}}M{2.8cm}|M{2.0cm}M{2.0cm}M{2.0cm}M{2.0cm}M{2.0cm}@{\hspace{0.1cm}}}
\toprule
\multirow{2}{*}[-0.1cm]{Method} & \multicolumn{5}{c}{Fraction of the Cityscapes Dataset} \\ \cmidrule(lr){2-6}
 & 0.3\% & 1\% & 3\% & 10\% & 100\% \\
\drule
Baseline & 41.83 & 49.15 & 59.07 & 69.02 & 79.40 \\ \midrule
InstructPix2Pix & 41.94 {\small\color{red}(+0.11)} & 48.17 {\small\color{blue}(-0.98)} & 60.43 {\small\color{red}(+1.36)} & 66.21 {\small\color{blue}(-2.81)} & 78.06 {\small\color{blue}(-1.34)} \\
DatasetDM & 42.82 {\small\color{red}(+0.99)} & 49.71 {\small\color{red}(+0.56)} & 60.31 {\small\color{red}(+1.24)} & 69.04 {\small\color{red}(+0.02)} & \underline{80.45} {\small\color{red}(+1.05)} \\
LoRA & 42.97 {\small\color{red}(+1.14)} & \underline{51.80} {\small\color{red}(+2.65)} & 60.22 {\small\color{red}(+1.15)} & \underline{69.21} {\small\color{red}(+0.19)} & 79.75 {\small\color{red}(+0.35)} \\
AdaLoRA & \underline{43.67} {\small\color{red}(+1.84)} & 48.21 {\small\color{blue}(-0.94)} & \underline{60.93} {\small\color{red}(+1.86)} & 68.32 {\small\color{blue}(-0.70)} & 78.62 {\small\color{blue}(-0.78)} \\
CA-LoRA (Ours) & \textbf{44.13 {\small\color{red}(+2.30)}} & \textbf{51.90 {\small\color{red}(+2.75)}} & \textbf{61.29 {\small\color{red}(+2.22)}} & \textbf{70.29 {\small\color{red}(+1.27)}} & \textbf{80.74 {\small\color{red}(+1.34)}} \\
\bottomrule
\end{tabular}
}
\vspace{-5pt}
\end{table*}

\begin{table*}[t]
\centering
\caption{
\textbf{Comparison of generated datasets for domain generalization (DG) of urban-scene segmentation (mIoU)}.
The experiments are conducted using various DG methods~\cite{xie2021segformer, hoyer2022daformer, hoyer2022hrda}.
Each first row presents the baseline model trained solely on the real dataset.
}
\label{tab:exp_dg}
\vspace{-5pt}
\resizebox{0.78\linewidth}{!}{
\begin{tabular}{@{\hspace{0.1cm}}M{2.1cm}|M{2.8cm}|M{1.4cm}M{1.2cm}M{1.2cm}M{1.3cm}|M{2.1cm}@{\hspace{0.1cm}}}

\toprule
DG Method & Method & ACDC & DZ & BDD & MV & Average \\
\drule
 & Baseline & 53.12 & 25.69 & 53.00 & 59.81 & 47.91 \\ \cmidrule(lr){2-7}
 & InstructPix2Pix & \underline{56.02} & 26.92 & 54.03 & 60.44 & 49.35 \small{\color{red} (+1.44)} \\
 & DATUM & 54.47 & \underline{29.25} & 53.19 & 61.22 & 49.53 \small{\color{red} (+1.61)} \\
ColorAug & DatasetDM & 53.80 & 27.70 & 53.54 & 60.75 & 48.95 \small{\color{red} (+1.04)} \\
 & LoRA & 54.25 & 28.42 & \underline{54.34} & \underline{61.42} & 49.61 \small{\color{red} (+1.70)} \\
 & AdaLoRA & 54.39 & 28.65 & 53.78 & \textbf{61.78} & \underline{49.65} \small{\color{red} (+1.74)} \\
 & CA-LoRA (Ours) & \textbf{56.07} & \textbf{29.75} & \textbf{54.35} & 61.40 & \textbf{50.39 \small{\color{red} (+2.49)}} \\
\midrule
 & Baseline & 53.98 & 27.82 & 54.29 & 62.69 & 49.70 \\ \cmidrule(lr){2-7}
 & InstructPix2Pix & 55.13 & 26.93 & 54.61 & 62.36 & 49.76 \small{\color{red} (+0.06)} \\
 & DATUM & 54.06 & 27.10 & 54.74 & 62.40 & 49.58 \small{\color{blue} (-0.12)} \\
DAFormer & DatasetDM & \underline{55.24} & 28.44 & 54.40 & \underline{63.18} & 50.32 \small{\color{red} (+0.62)} \\
 & LoRA & 54.64 & \underline{30.22} & \textbf{55.44} & \textbf{63.39} & \underline{50.92} \small{\color{red} (+1.22)} \\
 & AdaLoRA & 54.94 & 30.06 & \underline{55.41} & 63.10 & 50.88 \small{\color{red} (+1.18)} \\
 & CA-LoRA (Ours) & \textbf{55.83} & \textbf{31.68} & 54.68 & 63.09 & \textbf{51.32 \small{\color{red} (+1.63)}} \\
\midrule
 & Baseline & 58.48 & 29.46 & 56.12 & 64.27 & 52.08 \\ \cmidrule(lr){2-7}
 & InstructPix2Pix & \underline{58.50} & 29.56 & 56.10 & 64.10 & 52.07 \small{\color{blue} (-0.01)} \\
 & DATUM & 58.11 & 30.18 & \underline{56.94} & 64.29 & 52.38 \small{\color{red} (+0.30)} \\
HRDA & DatasetDM & 58.11 & 31.51 & 55.74 & 64.49 & 52.46 \small{\color{red} (+0.38)} \\
 & LoRA & 57.47 & 31.09 & \textbf{57.29} & 64.47 & 52.58 \small{\color{red} (+0.50)} \\
 & AdaLoRA & 58.37 & \underline{32.23} & 56.40 & \textbf{64.58} & \underline{52.90} \small{\color{red} (+0.82)} \\
 & CA-LoRA (Ours) & \textbf{58.93} & \textbf{34.41} & 56.56 & \underline{64.54} & \textbf{53.61 \small{\color{red} (+1.53)}} \\
\bottomrule

\end{tabular}
} 
\vspace{-0.5cm}
\end{table*}

\vspace{-0.2cm}
\section{Experiments}
\label{sec:4_experiments}

In the following sections, we present the answers to the following research questions:
\begin{enumerate}
\item How does the generated dataset contribute to improving segmentors across various settings? (\Cref{sec:4_exp_main})
\item How important is image-domain alignment in enhancing segmentation performance? (\Cref{sec:4_exp_main,sec:4_exp_different_concept})
\item How does the T2I model’s behavior change when the target concept is modified? (\Cref{sec:4_exp_different_concept})
\item How does concept-aware layer selection outperform hand-crafted baselines? (\Cref{sec:4_exp_hand_crafted_ablation})
\end{enumerate}


\begin{figure*}[t]
\centering
\includegraphics[width=\linewidth]{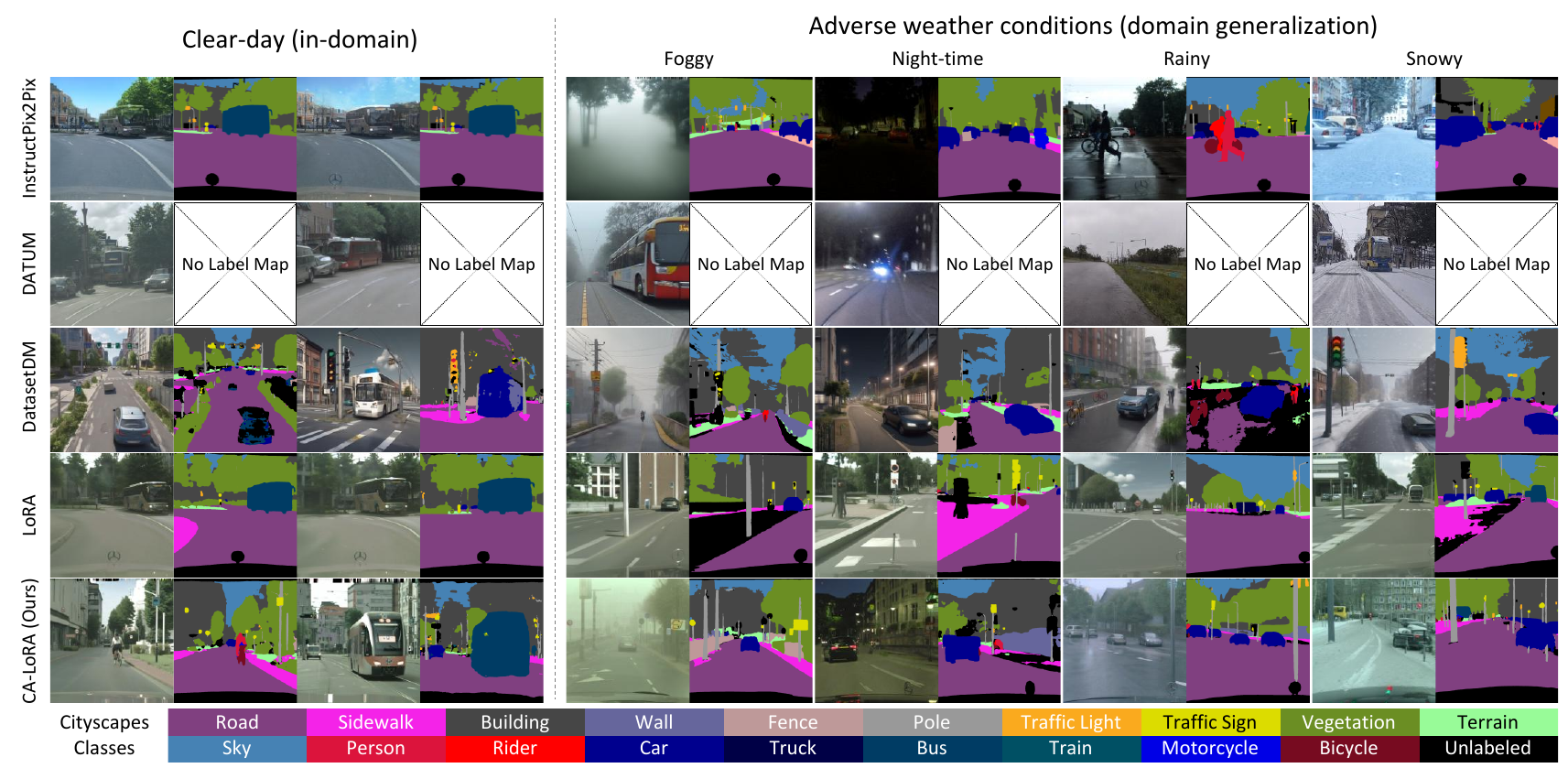}
\vspace{-15pt}
\caption{
\textbf{Qualitative comparison of image-label pairs.}
InstructPix2Pix only edits textures and thus cannot produce diverse scenes, while DATUM generates personalized but unlabeled images.
DatasetDM suffers from viewpoint and style misalignment due to the lack of adaptation, and LoRA tends to memorize training examples.
In contrast, CA-LoRA selectively learns style or viewpoint concepts, producing more diverse and better-aligned image-label pairs.
}
\label{fig:4_exp_quali_comparison}
\vspace{-15pt}
\end{figure*}

\subsection{Experimental Setup}
\label{sec:4_exp_setup}


\paragraph{Implementation Details}
We use SDXL~\cite{sdxl} as the pretrained T2I model.
For a fair comparison, we reimplement DatasetDM~\cite{datasetdm} with SDXL as well.
We fix the rank of both the original LoRA and CA-LoRA at 64 and set the training iterations to 10k.
We utilize the Style CA-LoRA for in-domain segmentation whereas Viewpoint CA-LoRA for domain generalization.
We explore the selected proportion of CA-LoRA across 1\%, 2\%, 3\%, 5\%, and 10\%.
The diffusion timestep $t$ for identifying the desired concept has been searched in [1, 81, 201, 481] out of 1000 timesteps~\cite{xu2023odise} and set to 81.
Although our method requires additional CA-LoRA fine-tuning, it takes only about one hour on a single V100 GPU, adding negligible overhead compared to the 20-hour label generator training.
The results of our hyperparameter search, along with details such as the label generator architecture, are provided in \Cref{appn:analysis_concept_awareness,appn:analysis_ca_lora,appn:imple}.

\paragraph{Baselines}
We compare our approach with prior dataset generation methods, including InstructPix2Pix~\cite{brooks2023instructpix2pix}, DATUM~\cite{benigmim2023one}\footnote{For DATUM~\cite{benigmim2023one}, we add one extra image per weather condition to meet the one-shot UDA setup and exclude it from the in-domain baseline, as it requires a UDA method.}, and DatasetDM~\cite{datasetdm}.
For fine-tuning approaches, we further compare our concept-aware method with LoRA~\cite{hu2022lora} and AdaLoRA~\cite{zhang2023adalora} to demonstrate the benefit of selective fine-tuning.
All fine-tuning methods are implemented on top of the DatasetDM framework.

\paragraph{In-Domain Semantic Segmentation}
For baseline model, we train Mask2Former~\cite{mask2former} on subsets of the Cityscapes~\cite{cityscapes} dataset at various fractions (0.3\%, 1\%, 3\%, 10\%, and 100\%), which we denote as `Baseline'.
Then, we generate 500 image-label pairs for all few-shot settings and use them as an additional dataset to fine-tune the baseline model, where we generate 3000 pairs for the fully-supervised setting.
To prevent overfitting to the generated data, we mix real (\eg, samples from the Cityscapes dataset) and generated samples in equal proportions within each mini-batch.

\paragraph{Domain Generalization in Semantic Segmentation}
For the evaluation, we train on the Cityscapes~\cite{cityscapes} and test on ACDC~\cite{acdc}, Dark Zurich (DZ)~\cite{dz}, BDD100K (BDD)~\cite{bdd100k}, and Mapillary Vistas (MV)~\cite{neuhold2017mapillary}.
We improve the DG performance for urban-scene segmentation by building upon existing DG methods, including ColorAug, DAFormer~\cite{hoyer2022daformer} and HRDA~\cite{hoyer2022hrda}. 
To show the effectiveness of the generated dataset, we train a semantic segmentation model with each DG method on a combination of the 2975 Cityscapes image-label pairs and the 2500 generated image-label pairs from scratch.
We generate 500 images for 5 adverse weather or lighting conditions including clear, foggy, night-time, rainy, and snowy.

\subsection{Main Results on the Segmentation Benchmarks}
\label{sec:4_exp_main}
\vspace{-5pt}


\paragraph{In-Domain Semantic Segmentation}
\Cref{tab:exp_cityscapes} shows that the proposed CA-LoRA consistently outperforms all other methods across various data ratios.
It improves mIoU by 2.30 at the 0.3\% data fraction of Cityscapes, significantly surpassing DatasetDM, which yields only a 0.99 increase.
Although the fine-tuning baselines (LoRA and AdaLoRA) show competitive performance at lower data fractions, they lag behind CA-LoRA at larger fractions (e.g., 10\%, 100\%) and struggle to generate informative samples beyond the source domain.
As shown in \Cref{fig:4_exp_quali_comparison}, InstructPix2Pix offers limited improvements because it mainly alters the texture of the scene and cannot meaningfully modify structural content.
The consistent gains across data ratios highlight the robustness of CA-LoRA, demonstrating its effectiveness for urban-scene segmentation in both few-shot and fully supervised settings.
We also observe consistent gains on PASCAL VOC~\cite{everingham2010pascal}, showing that CA-LoRA is applicable beyond urban-scene segmentation (see \Cref{appn:voc}).

\paragraph{Domain Generalization in Semantic Segmentation}
As shown in \Cref{tab:exp_dg}, the proposed method consistently outperforms existing segmentation dataset generation approaches across multiple DG methods~\cite{xie2021segformer,hoyer2022daformer,hoyer2022hrda}.
\Cref{fig:4_exp_quali_comparison} further illustrates that CA-LoRA learns only viewpoint information from the source dataset (Cityscapes) while preserving the pretrained T2I model’s diverse style-generation capability.
This results in strong generalization performance, particularly on challenging datasets such as ACDC and Dark Zurich, where adverse weather and lighting conditions drive the domain shift.
In contrast, BDD100K and Mapillary Vistas differ mainly in scene mood or appearance, leading to milder shifts compared to ACDC and DZ.
Finally, our approach improves performance not only over simple DG methods (ColorAug) but also when combined with advanced ones (DAFormer and HRDA), as our generated datasets provide semantically diverse content that label-to-image methods such as InstructPix2Pix and DGInStyle~\cite{jia2023dginstyle} struggle to reproduce.

\subsection{Style-aware LoRA vs. Viewpoint-aware LoRA}
\label{sec:4_exp_different_concept}

This section provides an in-depth analysis of the CA-LoRA fine-tuned models (style and viewpoint), comparing them with the pretrained T2I model (0\%, no fine-tuning; DatasetDM) and the LoRA fine-tuned model (100\%, fine-tuning all layers; LoRA) used in the main results.


%
%

\begin{figure*}
\centering
\includegraphics[width=0.9\linewidth]{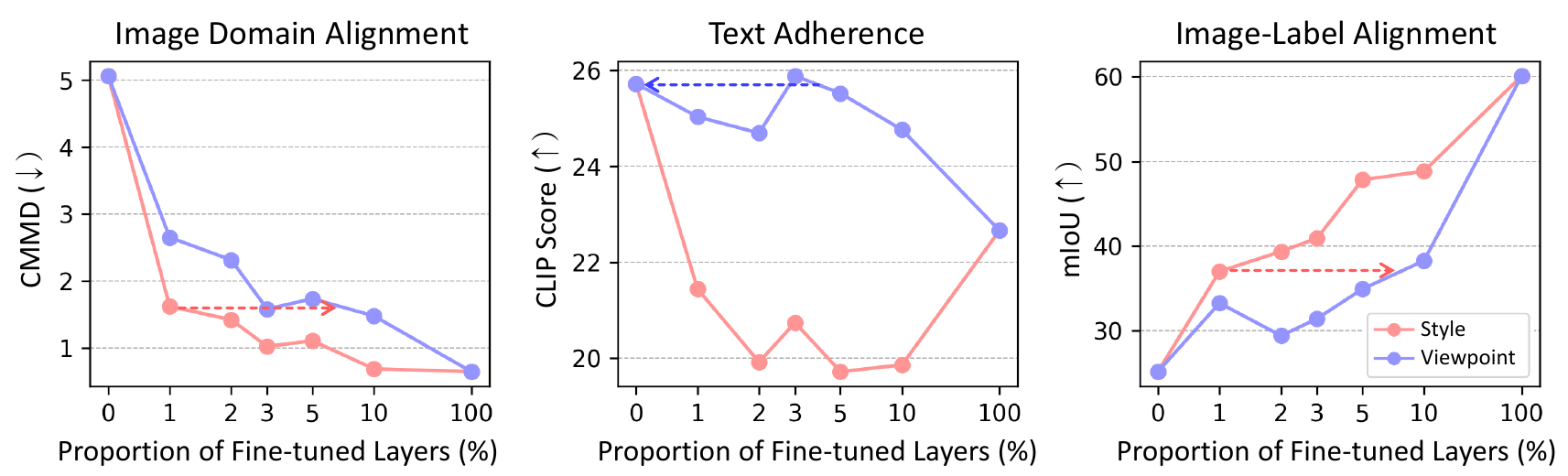}
\vspace{-5pt}
\caption{
\textbf{Comparison of Style CA-LoRA and Viewpoint CA-LoRA.}
Even when fine-tuning only 1\% of the style-aware layers, image-domain alignment improves as quickly as fine-tuning 5–10\% of the viewpoint-aware layers.
In contrast, viewpoint CA-LoRA maintains strong text adherence even when fine-tuning 3–5\% of its layers.
Finally, style-aware layers show an advantage in image-label alignment, as their generative features exhibit a smaller domain gap.
}
\label{fig:4_exp_analysis_cmmd_clip_miou}
\vspace{-12pt}
\end{figure*}

\paragraph{Image Domain Alignment}
We evaluate the image-domain alignment of T2I models, which is crucial for generating an in-domain dataset.
To measure this alignment, we adopt CMMD~\cite{cmmd}, as maximum mean discrepancy provides a reliable comparison for small image sets.
\Cref{fig:4_exp_analysis_cmmd_clip_miou} (left) shows that the pretrained T2I model exhibits a large domain gap between real and generated images, whereas fine-tuning on Cityscapes effectively reduces this gap.
Between the two CA-LoRA variants, Style CA-LoRA achieves comparable alignment while using only 10\% of the parameters of the full LoRA.
Moreover, fine-tuning just 1\% of the style-aware layers yields image-domain alignment similar to fine-tuning 5–10\% of the viewpoint-aware layers.
Although the full LoRA (100\%) achieves the best CMMD score, \Cref{fig:1_intro_motivation} shows that it suffers from severe memorization, providing limited useful variation and resulting in only marginal segmentation improvements.

\paragraph{Image Generation for Various Weather Scenarios}
Since generating diverse weather conditions (\eg, foggy, rainy) is crucial for improving DG performance, preserving the text adherence ability of the T2I model is essential.
We measure the weather-conditional image generation performance by leveraging CLIP-Score~\cite{clip}, which can assess the similarity between the generated images and their input text prompts.
\Cref{fig:4_exp_analysis_cmmd_clip_miou} shows the average CLIP-Score across four diverse weather conditions, based on 100 generated images per condition.
The results indicate that the pretrained model achieves a high CLIP score for generating adverse weather conditions, whereas the Style CA-LoRA drastically fails to do so, as they primarily learn styles from the source dataset, which inherently includes its weather condition (\eg, clear-day).
In contrast, the Viewpoint CA-LoRA effectively preserves the adverse weather conditional generation performance while learning the viewpoint from the source dataset.

\paragraph{Image-Label Alignment}
Generating reliable labels is a key requirement for constructing high-quality segmentation datasets.
As shown in \Cref{fig:4_exp_quali_comparison}, CA-LoRA achieves superior image–label alignment compared to the baselines by fine-tuning the T2I model for domain alignment.
To quantitatively evaluate this alignment, we obtain pseudo–ground truth for the generated images using predictions from strong pretrained segmentors~\cite{mask2former}.\footnote{This is reasonable when the pretrained segmentors are trained on substantially larger datasets than the label generator.}
As shown in \Cref{fig:4_exp_analysis_cmmd_clip_miou}, Style CA-LoRA achieves better image–label alignment than Viewpoint CA-LoRA, and increasing the number of fine-tuned layers further improves alignment.
This improvement arises from the reduced domain gap between the training and inference-time generative features of the T2I model.
However, increasing the number of fine-tuned layers is not always beneficial, as it can reduce scene diversity and produce less informative samples, ultimately yielding only marginal segmentation improvements.

\begin{table}[t]
\centering
\caption{
\textbf{Ablation on parameter group selection for fine-tuning.}
The image domain alignment is measured by CMMD, and the final segmentation performance (mIoU, Cityscapes 0.3\%).
}
\vspace{-8pt}
\resizebox{\linewidth}{!}{
\begin{tabular}{lccc}
\toprule
Finetuning Group & \% Params. & CMMD ($\downarrow$) & Performance ($\uparrow$) \\
\drule
No Finetuning (DatasetDM) & 0\% & 5.063 & 42.82 \\ \midrule
Q projections & 25\% & 4.305 & 40.50 {\small\color{blue}(-2.32)} \\
K projections & 25\% & 3.990 & \underline{43.50} {\small\color{red}(+0.68)} \\
V projections & 25\% & 3.003 & 42.77 {\small\color{blue}(-0.05)} \\
OUT projections & 25\% & 3.005 & 42.82 {\small(+0.00)} \\
All projections (LoRA) & 100\% & \textbf{0.644} & 42.97 {\small\color{red}(+0.15)} \\
Random selection & 2\% & 0.783 & 43.24 {\small\color{red}(+0.42)} \\
Concept-Awareness w/o norm. & 2\% & \underline{0.730} & 43.01 {\small\color{red}(+0.19)} \\
Concept-Awareness (Ours) & 2\% & 1.420 & \textbf{44.13 {\small\color{red}(+1.31)}} \\
\bottomrule
\end{tabular}
}
\label{tab:ablation_parameter_group}
\vspace{-10pt}
\end{table}

\subsection{Performance According to the Fine-tuning Parameter Group}
\label{sec:4_exp_hand_crafted_ablation}

To demonstrate the effectiveness of our parameter-selection approach, we conduct several ablation studies in the in-domain few-shot setting (Cityscapes 0.3\%).
We report improvements in both image-domain alignment and segmentation performance over the no-finetuning baseline (DatasetDM).
Our evaluation includes hand-crafted projection–layer selection strategies (Q, K, V, OUT), random selection, and Concept-Awareness–based selection without diffusion-gradient normalization (see \Cref{eq:3_method_awareness}).

As shown in \Cref{tab:ablation_parameter_group}, hand-crafted layer-selection strategies (Q, V, OUT) do not meaningfully improve either image-domain alignment or segmentation performance.
Among them, only K projections show noticeable gains, yet they still fall short of our method.
Fine-tuning all projection layers provides the best domain alignment, but it fails to generate informative samples due to memorization.
Selecting a random subset of projection layers alleviates this issue to some extent but remains limited, underscoring the need for concept-aware selection.
Concept-Awareness without normalization behaves similarly to full-layer fine-tuning, as it selects weights dominated by diffusion gradients; while parameter-efficient, it lacks true concept specificity.
In contrast, our Concept-Awareness method achieves competitive domain alignment and yields the largest segmentation improvement (+1.31 mIoU over DatasetDM), demonstrating the effectiveness of concept-aware fine-tuning.


\vspace{-0.12cm}
\section{Conclusion and Future Work}
\vspace{-0.11cm}

This paper proposes Concept-Aware LoRA (CA-LoRA), a novel fine-tuning method that learns only the desired concepts (\eg, viewpoint or style) from the training dataset to generate segmentation datasets. By selectively identifying and updating only the weights relevant to the desired concepts, CA-LoRA enables the fine-tuned T2I model to produce well-aligned and informative samples. Moreover, it effectively leverages pretrained knowledge from T2I models to generate datasets, significantly improving segmentation performance across various settings, including in-domain (few-shot and fully-supervised) and domain generalization tasks. 
Furthermore, CA-LoRA excels at learning only the desired concepts from training data, even when it contains diverse information, including unnecessary concepts for dataset generation, making it applicable to various knowledge transfer and fine-tuning tasks.  

\paragraph{Limitations}
While we carefully construct synthetic segmentation datasets that achieve state-of-the-art performance among dataset generation approaches, the resulting performance gains may still appear marginal in certain settings.
Nevertheless, adapting pretrained T2I models to better align with the target domain is crucial for dataset generation, and we believe our work provides a valuable foundation for principled approaches.
Furthermore, since our method primarily focuses on incorporating style and viewpoint concepts that are informative for urban-scene segmentation, extending the concept-awareness framework to capture other types of concepts beyond style and viewpoint represents a promising direction for future research.

{
    \small
    \bibliographystyle{ieeenat_fullname}
    \bibliography{main}

@String(CVPR= {IEEE Conf. Comput. Vis. Pattern Recog.})

@String(AAAI = {AAAI})

@String(CVPR  = {CVPR})

@article{datasetdm,
  title={Datasetdm: Synthesizing data with perception annotations using diffusion models},
  author={Wu, Weijia and Zhao, Yuzhong and Chen, Hao and Gu, Yuchao and Zhao, Rui and He, Yefei and Zhou, Hong and Shou, Mike Zheng and Shen, Chunhua},
  journal={Advances in Neural Information Processing Systems},
  volume={36},
  pages={54683--54695},
  year={2023}
}

@inproceedings{park2023learning,
  title={Learning to generate semantic layouts for higher text-image correspondence in text-to-image synthesis},
  author={Park, Minho and Yun, Jooyeol and Choi, Seunghwan and Choo, Jaegul},
  booktitle={Proceedings of the IEEE/CVF International Conference on Computer Vision},
  pages={7591--7600},
  year={2023}
}

@inproceedings{jia2023dginstyle,
  title={DGInStyle: Domain-Generalizable Semantic Segmentation with Image Diffusion Models and Stylized Semantic Control},
  author={Jia, Yuru and Hoyer, Lukas and Huang, Shengyu and Wang, Tianfu and Van Gool, Luc and Schindler, Konrad and Obukhov, Anton},
  booktitle={Synthetic Data for Computer Vision Workshop@ CVPR 2024},
  year={2023}
}

@inproceedings{wu2023diffumask,
  title={Diffumask: Synthesizing images with pixel-level annotations for semantic segmentation using diffusion models},
  author={Wu, Weijia and Zhao, Yuzhong and Shou, Mike Zheng and Zhou, Hong and Shen, Chunhua},
  booktitle={Proceedings of the IEEE/CVF International Conference on Computer Vision},
  pages={1206--1217},
  year={2023}
}

@article{khani2023slime,
  title={Slime: Segment like me},
  author={Khani, Aliasghar and Taghanaki, Saeid Asgari and Sanghi, Aditya and Amiri, Ali Mahdavi and Hamarneh, Ghassan},
  journal={arXiv preprint arXiv:2309.03179},
  year={2023}
}

@article{nguyen2024dataset_diffusion,
  title={Dataset diffusion: Diffusion-based synthetic data generation for pixel-level semantic segmentation},
  author={Nguyen, Quang and Vu, Truong and Tran, Anh and Nguyen, Khoi},
  journal={Advances in Neural Information Processing Systems},
  volume={36},
  year={2024}
}

@inproceedings{zhang2021datasetgan,
  title={Datasetgan: Efficient labeled data factory with minimal human effort},
  author={Zhang, Yuxuan and Ling, Huan and Gao, Jun and Yin, Kangxue and Lafleche, Jean-Francois and Barriuso, Adela and Torralba, Antonio and Fidler, Sanja},
  booktitle={Proceedings of the IEEE/CVF Conference on Computer Vision and Pattern Recognition},
  pages={10145--10155},
  year={2021}
}

@inproceedings{li2022bigdatasetgan,
  title={Bigdatasetgan: Synthesizing imagenet with pixel-wise annotations},
  author={Li, Daiqing and Ling, Huan and Kim, Seung Wook and Kreis, Karsten and Fidler, Sanja and Torralba, Antonio},
  booktitle={Proceedings of the IEEE/CVF Conference on Computer Vision and Pattern Recognition},
  pages={21330--21340},
  year={2022}
}

@inproceedings{baranchuk22label,
  title={Label-Efficient Semantic Segmentation with Diffusion Models},
  author={Baranchuk, Dmitry and Voynov, Andrey and Rubachev, Ivan and Khrulkov, Valentin and Babenko, Artem},
  booktitle={International Conference on Learning Representations},
  year={2022},
}

@article{yang2024freemask,
  title={Freemask: Synthetic images with dense annotations make stronger segmentation models},
  author={Yang, Lihe and Xu, Xiaogang and Kang, Bingyi and Shi, Yinghuan and Zhao, Hengshuang},
  journal={Advances in Neural Information Processing Systems},
  volume={36},
  year={2024}
}

@article{ddpm,
  title={Denoising diffusion probabilistic models},
  author={Ho, Jonathan and Jain, Ajay and Abbeel, Pieter},
  journal={Advances in neural information processing systems},
  volume={33},
  pages={6840--6851},
  year={2020}
}

@inproceedings{ldm,
  title={High-resolution image synthesis with latent diffusion models},
  author={Rombach, Robin and Blattmann, Andreas and Lorenz, Dominik and Esser, Patrick and Ommer, Bj{\"o}rn},
  booktitle={Proceedings of the IEEE/CVF conference on computer vision and pattern recognition},
  pages={10684--10695},
  year={2022}
}

@article{imagen,
  title={Photorealistic text-to-image diffusion models with deep language understanding},
  author={Saharia, Chitwan and Chan, William and Saxena, Saurabh and Li, Lala and Whang, Jay and Denton, Emily L and Ghasemipour, Kamyar and Gontijo Lopes, Raphael and Karagol Ayan, Burcu and Salimans, Tim and others},
  journal={Advances in Neural Information Processing Systems},
  volume={35},
  pages={36479--36494},
  year={2022}
}

@article{sdxl,
  title={Sdxl: Improving latent diffusion models for high-resolution image synthesis},
  author={Podell, Dustin and English, Zion and Lacey, Kyle and Blattmann, Andreas and Dockhorn, Tim and M{\"u}ller, Jonas and Penna, Joe and Rombach, Robin},
  journal={arXiv preprint arXiv:2307.01952},
  year={2023}
}

@article{dalle2,
  title={Hierarchical text-conditional image generation with clip latents},
  author={Ramesh, Aditya and Dhariwal, Prafulla and Nichol, Alex and Chu, Casey and Chen, Mark},
  journal={arXiv preprint arXiv:2204.06125},
  volume={1},
  number={2},
  pages={3},
  year={2022}
}

@inproceedings{sd3,
  title={Scaling rectified flow transformers for high-resolution image synthesis},
  author={Esser, Patrick and Kulal, Sumith and Blattmann, Andreas and Entezari, Rahim and M{\"u}ller, Jonas and Saini, Harry and Levi, Yam and Lorenz, Dominik and Sauer, Axel and Boesel, Frederic and others},
  booktitle={Forty-first International Conference on Machine Learning},
  year={2024}
}

@inproceedings{
    hu2022lora,
    title={Lo{RA}: Low-Rank Adaptation of Large Language Models},
    author={Edward J Hu and yelong shen and Phillip Wallis and Zeyuan Allen-Zhu and Yuanzhi Li and Shean Wang and Lu Wang and Weizhu Chen},
    booktitle={International Conference on Learning Representations},
    year={2022},
    url={https://openreview.net/forum?id=nZeVKeeFYf9}
}

@inproceedings{ruiz2023dreambooth,
  title={Dreambooth: Fine tuning text-to-image diffusion models for subject-driven generation},
  author={Ruiz, Nataniel and Li, Yuanzhen and Jampani, Varun and Pritch, Yael and Rubinstein, Michael and Aberman, Kfir},
  booktitle={Proceedings of the IEEE/CVF conference on computer vision and pattern recognition},
  pages={22500--22510},
  year={2023}
}

@article{laion5b,
  title={Laion-5b: An open large-scale dataset for training next generation image-text models},
  author={Schuhmann, Christoph and Beaumont, Romain and Vencu, Richard and Gordon, Cade and Wightman, Ross and Cherti, Mehdi and Coombes, Theo and Katta, Aarush and Mullis, Clayton and Wortsman, Mitchell and others},
  journal={Advances in Neural Information Processing Systems},
  volume={35},
  pages={25278--25294},
  year={2022}
}

@inproceedings{cmmd,
  title={Rethinking fid: Towards a better evaluation metric for image generation},
  author={Jayasumana, Sadeep and Ramalingam, Srikumar and Veit, Andreas and Glasner, Daniel and Chakrabarti, Ayan and Kumar, Sanjiv},
  booktitle={Proceedings of the IEEE/CVF Conference on Computer Vision and Pattern Recognition},
  pages={9307--9315},
  year={2024}
}

@inproceedings{cityscapes,
title={The Cityscapes Dataset for Semantic Urban Scene Understanding},
author={Cordts, Marius and Omran, Mohamed and Ramos, Sebastian and Rehfeld, Timo and Enzweiler, Markus and Benenson, Rodrigo and Franke, Uwe and Roth, Stefan and Schiele, Bernt},
booktitle={Proc. of the IEEE Conference on Computer Vision and Pattern Recognition (CVPR)},
year={2016}
}

@inproceedings{bdd100k,
  title={Bdd100k: A diverse driving dataset for heterogeneous multitask learning},
  author={Yu, Fisher and Chen, Haofeng and Wang, Xin and Xian, Wenqi and Chen, Yingying and Liu, Fangchen and Madhavan, Vashisht and Darrell, Trevor},
  booktitle={Proceedings of the IEEE/CVF conference on computer vision and pattern recognition},
  pages={2636--2645},
  year={2020}
}

@inproceedings{acdc,
  title={ACDC: The adverse conditions dataset with correspondences for semantic driving scene understanding},
  author={Sakaridis, Christos and Dai, Dengxin and Van Gool, Luc},
  booktitle={Proceedings of the IEEE/CVF International Conference on Computer Vision},
  pages={10765--10775},
  year={2021}
}

@inproceedings{dz,
  title={Guided curriculum model adaptation and uncertainty-aware evaluation for semantic nighttime image segmentation},
  author={Sakaridis, Christos and Dai, Dengxin and Gool, Luc Van},
  booktitle={Proceedings of the IEEE/CVF international conference on computer vision},
  pages={7374--7383},
  year={2019}
}

@inproceedings{neuhold2017mapillary,
  title={The mapillary vistas dataset for semantic understanding of street scenes},
  author={Neuhold, Gerhard and Ollmann, Tobias and Rota Bulo, Samuel and Kontschieder, Peter},
  booktitle={Proceedings of the IEEE international conference on computer vision},
  pages={4990--4999},
  year={2017}
}

@inproceedings{clip,
  title={Learning transferable visual models from natural language supervision},
  author={Radford, Alec and Kim, Jong Wook and Hallacy, Chris and Ramesh, Aditya and Goh, Gabriel and Agarwal, Sandhini and Sastry, Girish and Askell, Amanda and Mishkin, Pamela and Clark, Jack and others},
  booktitle={International conference on machine learning},
  pages={8748--8763},
  year={2021},
  organization={PMLR}
}

@article{shin2023fill,
  title={Fill-up: Balancing long-tailed data with generative models},
  author={Shin, Joonghyuk and Kang, Minguk and Park, Jaesik},
  journal={arXiv preprint arXiv:2306.07200},
  year={2023}
}

@inproceedings{xu2023odise,
  title={Open-vocabulary panoptic segmentation with text-to-image diffusion models},
  author={Xu, Jiarui and Liu, Sifei and Vahdat, Arash and Byeon, Wonmin and Wang, Xiaolong and De Mello, Shalini},
  booktitle={Proceedings of the IEEE/CVF Conference on Computer Vision and Pattern Recognition},
  pages={2955--2966},
  year={2023}
}

@article{kingma2014adam,
  title={Adam: A method for stochastic optimization},
  author={Kingma, Diederik P},
  journal={arXiv preprint arXiv:1412.6980},
  year={2014}
}

@inproceedings{adamw,
  title={Decoupled Weight Decay Regularization},
  author={Loshchilov, Ilya and Hutter, Frank},
  booktitle={International Conference on Learning Representations},
  year={2019}
}

@inproceedings{choi2021robustnet,
  title={Robustnet: Improving domain generalization in urban-scene segmentation via instance selective whitening},
  author={Choi, Sungha and Jung, Sanghun and Yun, Huiwon and Kim, Joanne T and Kim, Seungryong and Choo, Jaegul},
  booktitle={Proceedings of the IEEE/CVF conference on computer vision and pattern recognition},
  pages={11580--11590},
  year={2021}
}

@inproceedings{hoyer2022hrda,
  title={Hrda: Context-aware high-resolution domain-adaptive semantic segmentation},
  author={Hoyer, Lukas and Dai, Dengxin and Van Gool, Luc},
  booktitle={European conference on computer vision},
  pages={372--391},
  year={2022},
  organization={Springer}
}

@inproceedings{hoyer2022daformer,
  title={Daformer: Improving network architectures and training strategies for domain-adaptive semantic segmentation},
  author={Hoyer, Lukas and Dai, Dengxin and Van Gool, Luc},
  booktitle={Proceedings of the IEEE/CVF conference on computer vision and pattern recognition},
  pages={9924--9935},
  year={2022}
}

@inproceedings{mask2former,
  title={Masked-attention mask transformer for universal image segmentation},
  author={Cheng, Bowen and Misra, Ishan and Schwing, Alexander G and Kirillov, Alexander and Girdhar, Rohit},
  booktitle={Proceedings of the IEEE/CVF conference on computer vision and pattern recognition},
  pages={1290--1299},
  year={2022}
}

@article{wang2024instantstyle,
  title={Instantstyle: Free lunch towards style-preserving in text-to-image generation},
  author={Wang, Haofan and Wang, Qixun and Bai, Xu and Qin, Zekui and Chen, Anthony},
  journal={arXiv preprint arXiv:2404.02733},
  year={2024}
}

@article{xing2024csgo,
  title={CSGO: Content-Style Composition in Text-to-Image Generation},
  author={Xing, Peng and Wang, Haofan and Sun, Yanpeng and Wang, Qixun and Bai, Xu and Ai, Hao and Huang, Renyuan and Li, Zechao},
  journal={arXiv preprint arXiv:2408.16766},
  year={2024}
}

@article{zhang2023adalora,
  title={AdaLoRA: Adaptive budget allocation for parameter-efficient fine-tuning},
  author={Zhang, Qingru and Chen, Minshuo and Bukharin, Alexander and Karampatziakis, Nikos and He, Pengcheng and Cheng, Yu and Chen, Weizhu and Zhao, Tuo},
  journal={arXiv preprint arXiv:2303.10512},
  year={2023}
}

@article{liu2024dora,
  title={Dora: Weight-decomposed low-rank adaptation},
  author={Liu, Shih-Yang and Wang, Chien-Yi and Yin, Hongxu and Molchanov, Pavlo and Wang, Yu-Chiang Frank and Cheng, Kwang-Ting and Chen, Min-Hung},
  journal={arXiv preprint arXiv:2402.09353},
  year={2024}
}

@article{hayou2024lora+,
  title={Lora+: Efficient low rank adaptation of large models},
  author={Hayou, Soufiane and Ghosh, Nikhil and Yu, Bin},
  journal={arXiv preprint arXiv:2402.12354},
  year={2024}
}

@article{kopiczko2023vera,
  title={Vera: Vector-based random matrix adaptation},
  author={Kopiczko, Dawid Jan and Blankevoort, Tijmen and Asano, Yuki Markus},
  journal={arXiv preprint arXiv:2310.11454},
  year={2023}
}

@article{he2022sparseadapter,
  title={Sparseadapter: An easy approach for improving the parameter-efficiency of adapters},
  author={He, Shwai and Ding, Liang and Dong, Daize and Zhang, Miao and Tao, Dacheng},
  journal={arXiv preprint arXiv:2210.04284},
  year={2022}
}

@article{ding2023sparse,
  title={Sparse low-rank adaptation of pre-trained language models},
  author={Ding, Ning and Lv, Xingtai and Wang, Qiaosen and Chen, Yulin and Zhou, Bowen and Liu, Zhiyuan and Sun, Maosong},
  journal={arXiv preprint arXiv:2311.11696},
  year={2023}
}

@article{gal2022image,
  title={An image is worth one word: Personalizing text-to-image generation using textual inversion},
  author={Gal, Rinon and Alaluf, Yuval and Atzmon, Yuval and Patashnik, Or and Bermano, Amit H and Chechik, Gal and Cohen-Or, Daniel},
  journal={arXiv preprint arXiv:2208.01618},
  year={2022}
}

@inproceedings{basu2024mechanistic,
  title={On Mechanistic Knowledge Localization in Text-to-Image Generative Models},
  author={Basu, Samyadeep and Rezaei, Keivan and Kattakinda, Priyatham and Morariu, Vlad I and Zhao, Nanxuan and Rossi, Ryan A and Manjunatha, Varun and Feizi, Soheil},
  booktitle={Forty-first International Conference on Machine Learning},
  year={2024}
}

@inproceedings{tranheden2021dacs,
  title={Dacs: Domain adaptation via cross-domain mixed sampling},
  author={Tranheden, Wilhelm and Olsson, Viktor and Pinto, Juliano and Svensson, Lennart},
  booktitle={Proceedings of the IEEE/CVF winter conference on applications of computer vision},
  pages={1379--1389},
  year={2021}
}

@article{paszke2019pytorch,
  title={Pytorch: An imperative style, high-performance deep learning library},
  author={Paszke, Adam and Gross, Sam and Massa, Francisco and Lerer, Adam and Bradbury, James and Chanan, Gregory and Killeen, Trevor and Lin, Zeming and Gimelshein, Natalia and Antiga, Luca and others},
  journal={Advances in neural information processing systems},
  volume={32},
  year={2019}
}

@Misc{peft_github,
  title =        {PEFT: State-of-the-art Parameter-Efficient Fine-Tuning methods},
  author =       {Sourab Mangrulkar and Sylvain Gugger and Lysandre Debut and Younes Belkada and Sayak Paul and Benjamin Bossan},
  howpublished = {\url{https://github.com/huggingface/peft}},
  year =         {2022}
}

@misc{von-platen-etal-2022-diffusers,
  author = {Patrick von Platen and Suraj Patil and Anton Lozhkov and Pedro Cuenca and Nathan Lambert and Kashif Rasul and Mishig Davaadorj and Dhruv Nair and Sayak Paul and William Berman and Yiyi Xu and Steven Liu and Thomas Wolf},
  title = {Diffusers: State-of-the-art diffusion models},
  year = {2022},
  publisher = {GitHub},
  journal = {GitHub repository},
  howpublished = {\url{https://github.com/huggingface/diffusers}}
}

@inproceedings{guo2019spottune,
  title={Spottune: transfer learning through adaptive fine-tuning},
  author={Guo, Yunhui and Shi, Honghui and Kumar, Abhishek and Grauman, Kristen and Rosing, Tajana and Feris, Rogerio},
  booktitle=CVPR,
  year={2019}
}

@inproceedings{choi2022improving,
  title={Improving test-time adaptation via shift-agnostic weight regularization and nearest source prototypes},
  author={Choi, Sungha and Yang, Seunghan and Choi, Seokeon and Yun, Sungrack},
  booktitle={European Conference on Computer Vision},
  pages={440--458},
  year={2022},
  organization={Springer}
}

@article{lee2022surgical,
  title={Surgical fine-tuning improves adaptation to distribution shifts},
  author={Lee, Yoonho and Chen, Annie S and Tajwar, Fahim and Kumar, Ananya and Yao, Huaxiu and Liang, Percy and Finn, Chelsea},
  journal={arXiv preprint arXiv:2210.11466},
  year={2022}
}

@inproceedings{tied_lora,
    title = "Tied-{L}o{RA}: Enhancing parameter efficiency of {L}o{RA} with Weight Tying",
    author = "Renduchintala, Adithya  and
      Konuk, Tugrul  and
      Kuchaiev, Oleksii",
    booktitle = "Proceedings of the 2024 Conference of the North American Chapter of the Association for Computational Linguistics: Human Language Technologies (Volume 1: Long Papers)",
    month = jun,
    year = "2024",
}

@InProceedings{gs_lora,
    author    = {Zhao, Hongbo and Ni, Bolin and Fan, Junsong and Wang, Yuxi and Chen, Yuntao and Meng, Gaofeng and Zhang, Zhaoxiang},
    title     = {Continual Forgetting for Pre-trained Vision Models},
    booktitle = {Proceedings of the IEEE/CVF Conference on Computer Vision and Pattern Recognition (CVPR)},
    month     = {June},
    year      = {2024},
    pages     = {28631-28642}
}

@misc{lora_sp,
      title={LoRA-SP: Streamlined Partial Parameter Adaptation for Resource-Efficient Fine-Tuning of Large Language Models}, 
      author={Yichao Wu and Yafei Xiang and Shuning Huo and Yulu Gong and Penghao Liang},
      year={2024},
      archivePrefix={arXiv},
}

@inproceedings{brooks2023instructpix2pix,
  title={Instructpix2pix: Learning to follow image editing instructions},
  author={Brooks, Tim and Holynski, Aleksander and Efros, Alexei A},
  booktitle={Proceedings of the IEEE/CVF Conference on Computer Vision and Pattern Recognition},
  pages={18392--18402},
  year={2023}
}

@article{xie2021segformer,
  title={SegFormer: Simple and efficient design for semantic segmentation with transformers},
  author={Xie, Enze and Wang, Wenhai and Yu, Zhiding and Anandkumar, Anima and Alvarez, Jose M and Luo, Ping},
  journal={Advances in neural information processing systems},
  volume={34},
  pages={12077--12090},
  year={2021}
}

@article{everingham2010pascal,
  title={The pascal visual object classes (voc) challenge},
  author={Everingham, Mark and Van Gool, Luc and Williams, Christopher KI and Winn, John and Zisserman, Andrew},
  journal={International journal of computer vision},
  volume={88},
  pages={303--338},
  year={2010},
  publisher={Springer}
}

@inproceedings{benigmim2023one,
  title={One-shot unsupervised domain adaptation with personalized diffusion models},
  author={Benigmim, Yasser and Roy, Subhankar and Essid, Slim and Kalogeiton, Vicky and Lathuili{\`e}re, St{\'e}phane},
  booktitle={Proceedings of the IEEE/CVF conference on computer vision and pattern recognition},
  pages={698--708},
  year={2023}
}

@article{gong2023prompting,
  title={Prompting diffusion representations for cross-domain semantic segmentation},
  author={Gong, Rui and Danelljan, Martin and Sun, Han and Mangas, Julio Delgado and Van Gool, Luc},
  journal={arXiv preprint arXiv:2307.02138},
  year={2023}
}

@inproceedings{kim2024weakly,
  title={Weakly Supervised Semantic Segmentation for Driving Scenes},
  author={Kim, Dongseob and Lee, Seungho and Choe, Junsuk and Shim, Hyunjung},
  booktitle={Proceedings of the AAAI Conference on Artificial Intelligence},
  volume={38},
  number={3},
  pages={2741--2749},
  year={2024}
}

@article{kaplan2024domain,
  title={Domain-Aware Fine-Tuning of Foundation Models},
  author={Kaplan, Ugur Ali and Keuper, Margret and Khoreva, Anna and Zhang, Dan and Li, Yumeng},
  journal={arXiv preprint arXiv:2407.03482},
  year={2024}
}
}


\newpage
\clearpage

\appendix

\let\paragraph\origparagraph




\maketitlesupplementary

\section*{Appendix}
In the supplementary material, we provide additional analyses, experiments, qualitative and quantitative results, and implementation details, including pseudocode, due to space limitations in the main manuscript.
Specifically, we describe implementation details, including PyTorch-like pseudocode, to ensure reproducibility (\Cref{appn:imple}).
Next, we illustrate an implicit bias of concept gradients (\Cref{appn:analysis_concept_awareness}) and additional experimental studies of our design choice of CA-LoRA (\Cref{appn:analysis_ca_lora}).
We then demonstrate the generalizability of the proposed method using a general-domain dataset, Pascal VOC~\cite{everingham2010pascal} (\Cref{appn:voc}).
Finally, we discuss class-aware dataset generation, which could be a promising future research direction for handling long-tailed distributions by leveraging text prompts (\Cref{appn:class_specific_generation}).
Below is the table of contents for the supplementary material.

\startcontents[appendices]
\printcontents[appendices]{l}{1}{\setcounter{tocdepth}{2}}

\clearpage

\twocolumn[{%
\renewcommand\twocolumn[1][]{#1}%
\begin{center}
\centering
\captionsetup{type=figure}

\centering
\includegraphics[width=1.0\linewidth]{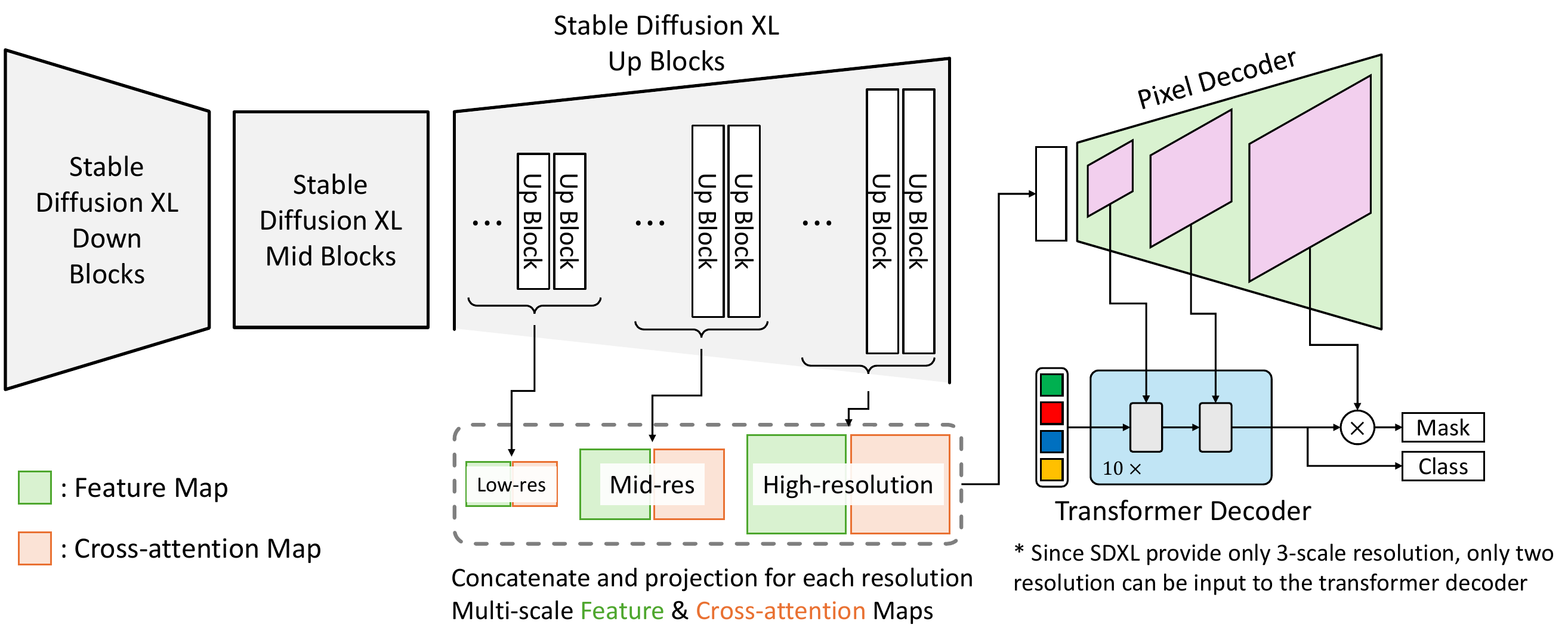}
\captionof{figure}{
The detailed label generator architecture. The whole framework includes a text-to-image generation model (Stable Diffusion XL), pixel decoder, and transformer decoder, followed by DatasetDM~\cite{datasetdm}. Due to the change in the architecture of the text-to-image generation model, the following pixel decoder and transformer decoder minorly changed (\eg, the number of input channels and the number of blocks).
}
\label{fig:A_detailed_label_generator}

\end{center}
}]

\section{Implementation Details}
\label{appn:imple}

\begin{figure*}[t]
\centering
\includegraphics[width=1.0\linewidth]{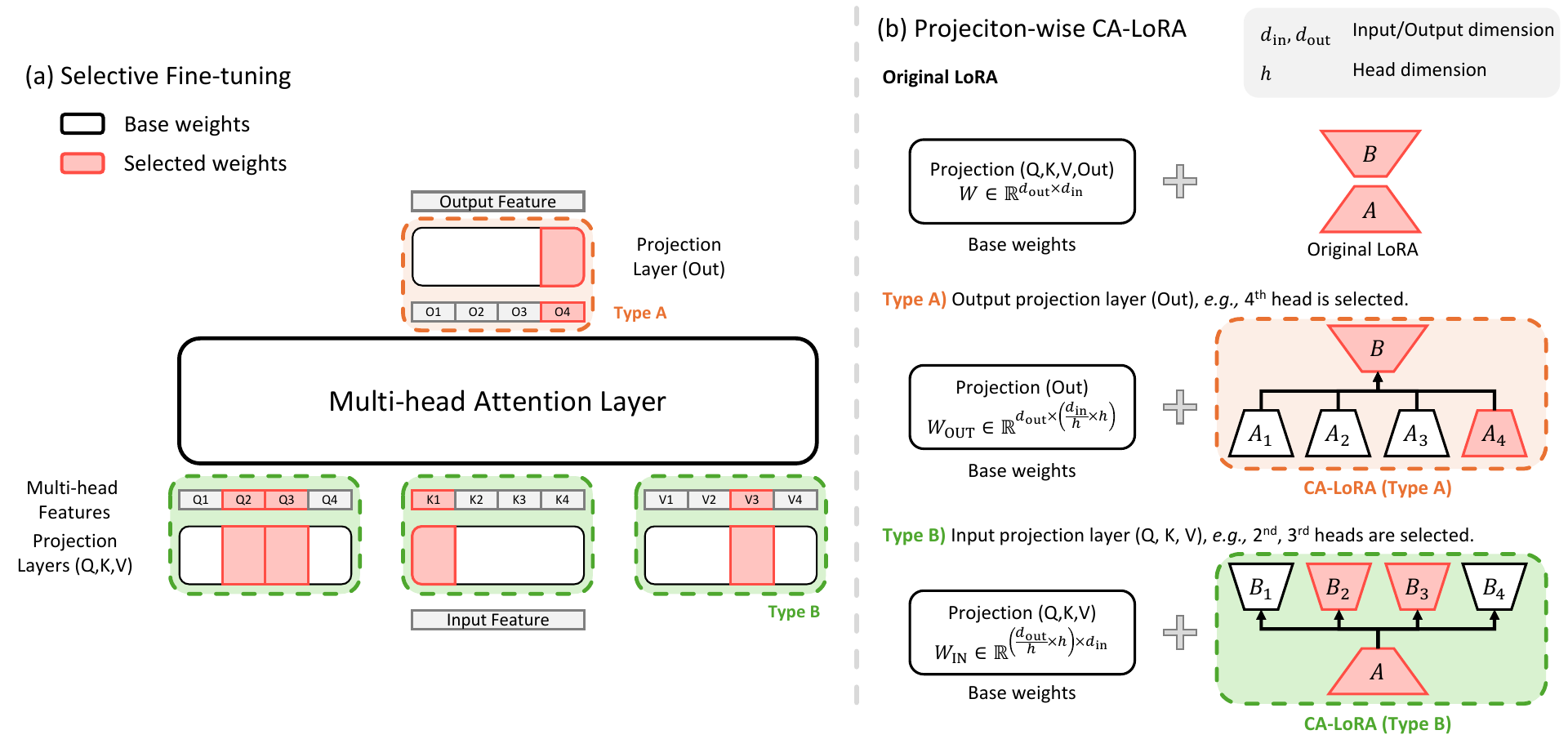}
\caption{The detailed architecture of the CA-LoRA. We conduct projection-wise CA-LoRA that can attach the LoRA layer for each projection layer of multi-head self-attention.}
\label{fig:A_method_slora_detailed}
\end{figure*}

\subsection{Label Generator Architecture}
We build the label generator based on the recent segmentation dataset generation framework, DatasetDM~\cite{datasetdm}.
The label generator in DatasetDM, called P-Decoder, is derived from the Mask2Former~\cite{mask2former} decoder architecture.
It takes intermediate features from the T2I model, including feature maps and cross-attention maps.
The label generator then concatenates features of the same resolution and reduces the feature dimensions using predefined projection layers.
The multi-resolution feature maps are passed through the pixel decoder and the transformer decoder sequentially, which outputs the segmentation predictions.
Finally, we calculate the loss function of the label decoder, which mirrors that of Mask2Former, incorporating binary cross-entropy, dice loss, and classification loss. 
However, several modifications exist between the original DatasetDM P-Decoder and our label generator due to architectural differences between Stable Diffusion v1.5~\cite{ldm} and Stable Diffusion XL~\cite{sdxl}.

Since DatasetDM is built on top of Stable Diffusion v1.5~\cite{ldm}, we simply adjust the feature dimensions in the projection layers to accommodate Stable Diffusion XL~\cite{sdxl}. The detailed label generator architecture is illustrated in \Cref{fig:A_detailed_label_generator}. 
However, if all feature maps and cross-attention maps are used, the total number of channels increases significantly, leading to an unmanageable number of parameters in the projection layers during concatenation and projection. 
In summary, the feature maps are extracted from the \textit{last feature block} at each resolution of the up-sampling blocks, while cross-attention maps are sampled at equal intervals (every 7 blocks) from the total 36 up-sampling blocks (i.e., 1st, 8th, ... 29th, 36th).

Furthermore, as shown in \Cref{fig:A_detailed_label_generator}, Stable Diffusion XL has only three resolution levels, compared to four resolution levels of the Stable Diffusion v1.5 architecture in the original DatasetDM.
In the Mask2Former structure, feature maps from the pixel decoder, excluding the largest resolution, are fed into the transformer decoder.
While the original design used three-resolution feature maps, only two were utilized in this case.
Thus, while DatasetDM provides three-resolution feature maps three times for 9 transformer decoder blocks, we provide two-resolution feature maps five times, leading to a total of 10 transformer decoder blocks. 
\emph{Importantly, to ensure a fair comparison, the reported scores for DatasetDM were obtained using a re-implemented version based on SDXL with the same modifications.}

\subsection{Detailed Architecture of CA-LoRA}
\label{appn:head_wise_lora}

\paragraph{Overview (\Cref{fig:A_method_slora_detailed} (a))}
The fundamental group of weights used to measure concept awareness is the linear projection layer.
We selectively adapt the pretrained weights layer by layer within the projection layers.
Practically, the projection layers in the multi-head attention block are integrated across heads, we split them to structurally distinguish the weights for each projection.
Consequently, we attach LoRA layers projection-wise, as shown in \Cref{fig:A_method_slora_detailed}.

\paragraph{Projection-wise CA-LoRA (\Cref{fig:A_method_slora_detailed} (b))}
There are two types of CA-LoRA: output LoRA projection layers (Type A: Output (OUT)) and input LoRA projection layers (Type B: Query (Q), Key (K), Value (V)).
To split the original LoRA layer ($\Delta W = B A$) in projection-wise, the output projection LoRA layers ($\Delta W_\text{OUT}$) split the $A$ weights row-wise, while the input projection LoRA layers ($\Delta W_\text{IN}$) split the $B$ weights column-wise, as illustrated in the following equations.

\begin{equation}
\hspace{-1cm}
\Delta W_\text{OUT} = B
\begin{bmatrix}
 &  & & \\
A_1 & A_2 & \cdots & A_h \\
 &  & &
\end{bmatrix},
\end{equation}

\begin{equation}
\Delta W_\text{IN} = 
\begin{bmatrix}
 & B_1 & \\
 & B_2 &  \\
 & \vdots & \\
 & B_h & 
\end{bmatrix}
A.
\end{equation}
The projection-wise LoRA is represented in \Cref{fig:A_method_slora_detailed} and \Cref{algo:ca_lora_define}.

\subsection{Hyperparameters and Pseudocode}
\label{appn:pseudocode}

\paragraph{Hyperparameters (\Cref{tab:A_hparam1,,tab:A_hparam2,,tab:A_hparam3,,tab:A_hparam4})}
We provide all hyperparameters to support reproducibility.
In the first stage, we fine-tune Stable Diffusion XL~\cite{sdxl} using the \href{https://huggingface.co/docs/diffusers/index}{HuggingFace Diffusers} library~\cite{von-platen-etal-2022-diffusers}. The specific hyperparameters for fine-tuning Stable Diffusion XL on the Cityscapes dataset are listed in \Cref{tab:A_hparam1}, while the training configurations for the label generator can be found in \Cref{tab:A_hparam2}.

Next, we train segmentation models for both in-domain and domain generalization scenarios. The hyperparameters for in-domain fine-tuning are provided in \Cref{tab:A_hparam3}, while those for domain generalization, based on the DGInStyle~\cite{jia2023dginstyle} method, are included in \Cref{tab:A_hparam4}. We hope these provided hyperparameters will facilitate reproducibility.

\begin{table}[t!]
\centering
\caption{\textbf{Hyperparameter search of layer proportions.} Image-domain alignment (CMMD) and final segmentation performance.}
\label{tab:vertical_ablation}
\resizebox{0.85\linewidth}{!}{
\begin{tabular}{lcc}

\toprule
Proportion & CMMD & Performance (mIoU) \\ 
\midrule
0\% (Pretrained) & 5.063 & 42.82 \\
1\% & 1.618 & 43.77 \\
2\% & 1.420 & \textbf{44.13} \\
3\% & 1.021 & \underline{43.94} \\
5\% & 1.105 & 43.81 \\
10\% & \underline{0.686} & 43.36 \\
100\% (Original LoRA) & \textbf{0.644} & 42.97 \\
\bottomrule

\end{tabular}
}
\vspace{-10pt}
\end{table}

\paragraph{Pseudocode}
We also provide PyTorch-like pseudocode~\cite{paszke2019pytorch} for key algorithms to effectively support reproducibility.
\textbf{Concept awareness (\Cref{algo:concept_awareness,algo:concept_helper})}
The concept awareness algorithm is demonstrated in \Cref{algo:concept_awareness}.
Conducting concept awareness requires several helper functions, as shown in \Cref{algo:concept_helper}.
\textbf{CA-LoRA (\Cref{algo:ca_lora_forward,algo:ca_lora_define})}
The CA-LoRA algorithm is divided into two parts: the forward function and the declaration function.
The forward pass of CA-LoRA is presented in \Cref{algo:ca_lora_forward}, while the declaration function, along with the selected layers, is illustrated in \Cref{algo:ca_lora_define}.

While we provide the PyTorch-like pseudocode based on HuggingFace Diffusers library~\cite{von-platen-etal-2022-diffusers}, HuggingFace PEFT-based implementation~\cite{peft_github} can reduce the training time of the CA-LoRA.

\clearpage

\twocolumn[{%
\renewcommand\twocolumn[1][]{#1}%
\begin{center}
\centering
\captionsetup{type=figure}

\includegraphics[width=\linewidth]{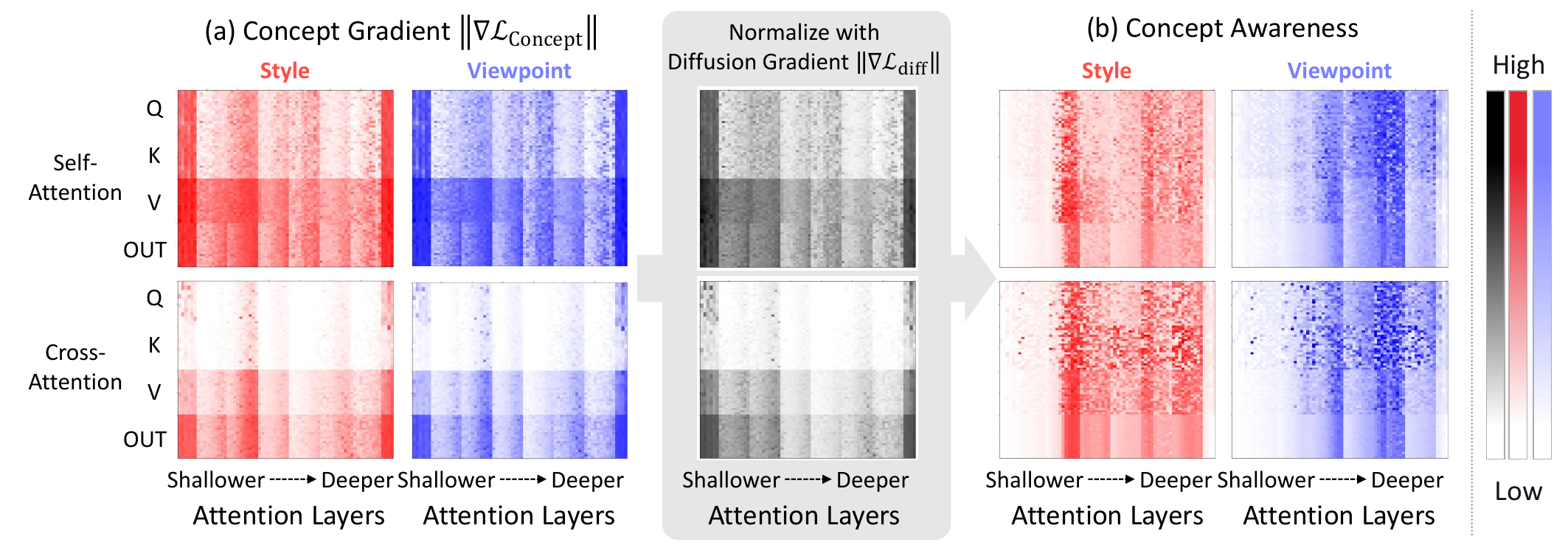}
\captionof{figure}{
\textbf{Visualizing the gradients and concept awareness.
}(a) Implicit gradient biases across network layers.
(b) Concept awareness, computed by normalizing each concept gradient with respect to the original diffusion-loss gradient.
Style-aware and viewpoint-aware weights are highlighted in red and blue, respectively.
Although raw concept gradients alone cannot clearly separate different concepts, normalization with diffusion gradients yields a more distinguishable concept-awareness signal.
}
\label{fig:A_gradient_bias}

\end{center}
}]

\section{Visualizing Concept Awareness}
\label{appn:analysis_concept_awareness}

\subsection{The Implicit Bias of Gradients Across Layers}
\label{appn:bias_of_gradients}

\paragraph{Observation (\Cref{fig:A_gradient_bias} (a))}
We compute the awareness scores for each layer by using the norm of the gradient. However, the gradient norm cannot be uniformly scaled across different head types (Q, K, V, OUT), attention types (self, cross), and layers (shallow, deep). To address this, we construct a base gradient to scale the concept loss gradient by referencing the gradient of the original diffusion loss, as described in \Cref{sec:3_method_measure_score}. We visualize the gradients of both the concept losses and the original diffusion loss in \Cref{fig:A_gradient_bias}. In this visualization, we separate the self-attention and cross-attention layers to provide clearer distinctions, which differs from the approach in the main paper.


\paragraph{Normalizing Gradients (\Cref{fig:A_gradient_bias} (b))}
Therefore, we normalize the gradients using the gradients calculated from the original diffusion loss, as discussed in \Cref{sec:3_method_measure_score} and shown in \Cref{fig:3_method_1_measure}.
As shown in \Cref{fig:A_gradient_bias} (a), the gradients calculated from the style concept loss and viewpoint concept loss are similar. However, the gradient increase ratio can differ significantly, as illustrated in \Cref{fig:A_gradient_bias} (b).

\begin{figure*}[t]
\centering
\includegraphics[width=1.0\linewidth]{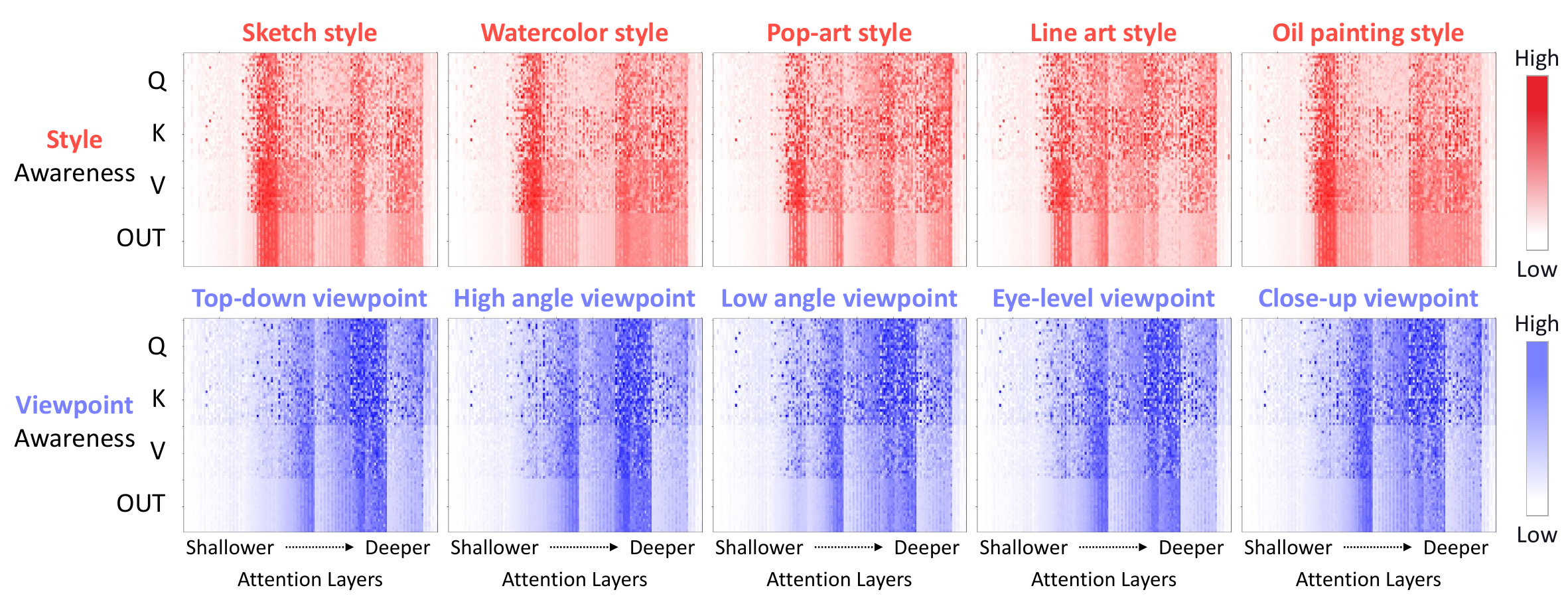}
\captionof{figure}{
\textbf{Measured concept awareness according to the various prompt augmentation.}
The highlighted concept-aware layers for each concept (style and viewpoint) remained largely consistent regardless of the prompt augmentation, demonstrating the robustness of concept awareness to variations in prompt design.
}
\label{fig:A_awareness_robustness}
\vspace{-5pt}
\end{figure*}

\begin{figure*}[t]
\centering
\includegraphics[width=0.85\linewidth]{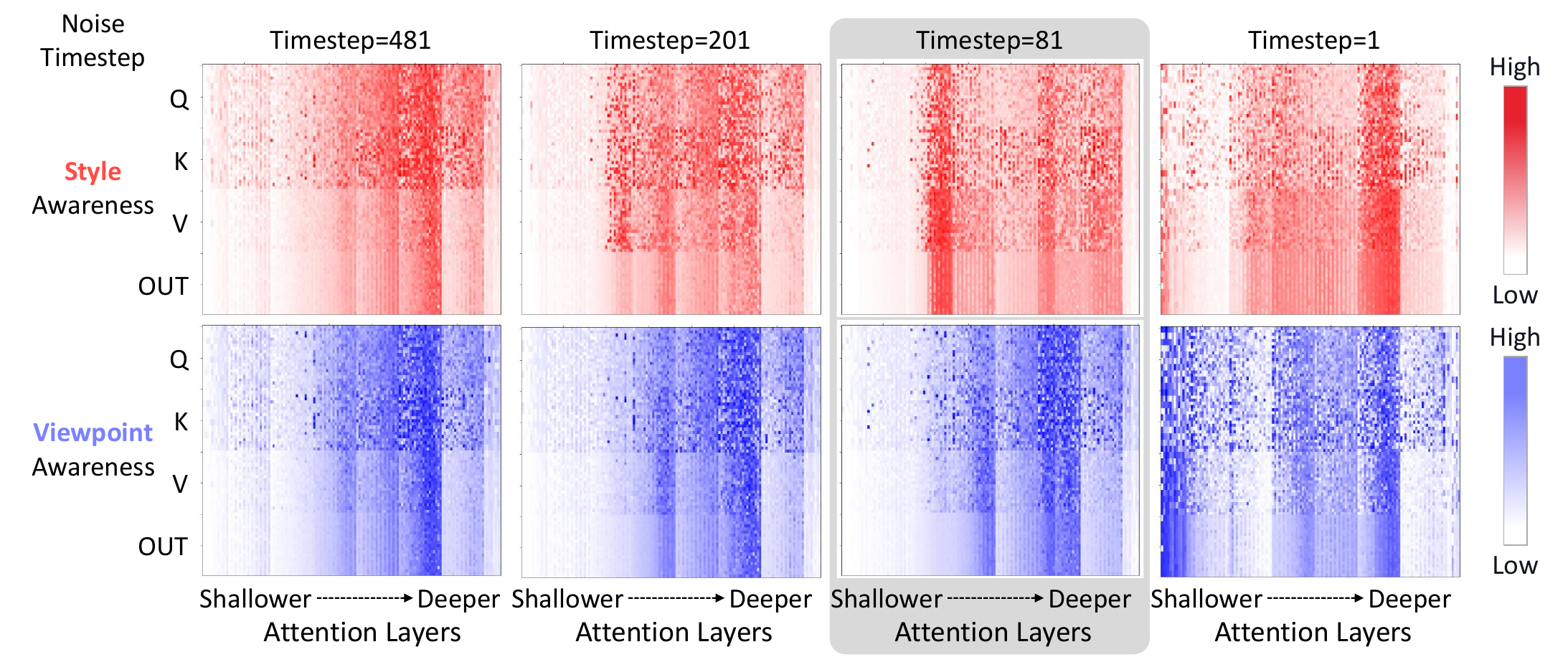}
\caption{
\textbf{Visualizing concept awareness across different noise timesteps.} It shows that the 81st timestep stands out with a significantly distinct concept awareness score between style and viewpoint awareness compared to the other timesteps.
}
\label{fig:A_concept_awareness_timestep}
\vspace{-5pt}
\end{figure*}

\begin{figure*}[t]
\centering
\includegraphics[width=0.85\linewidth]{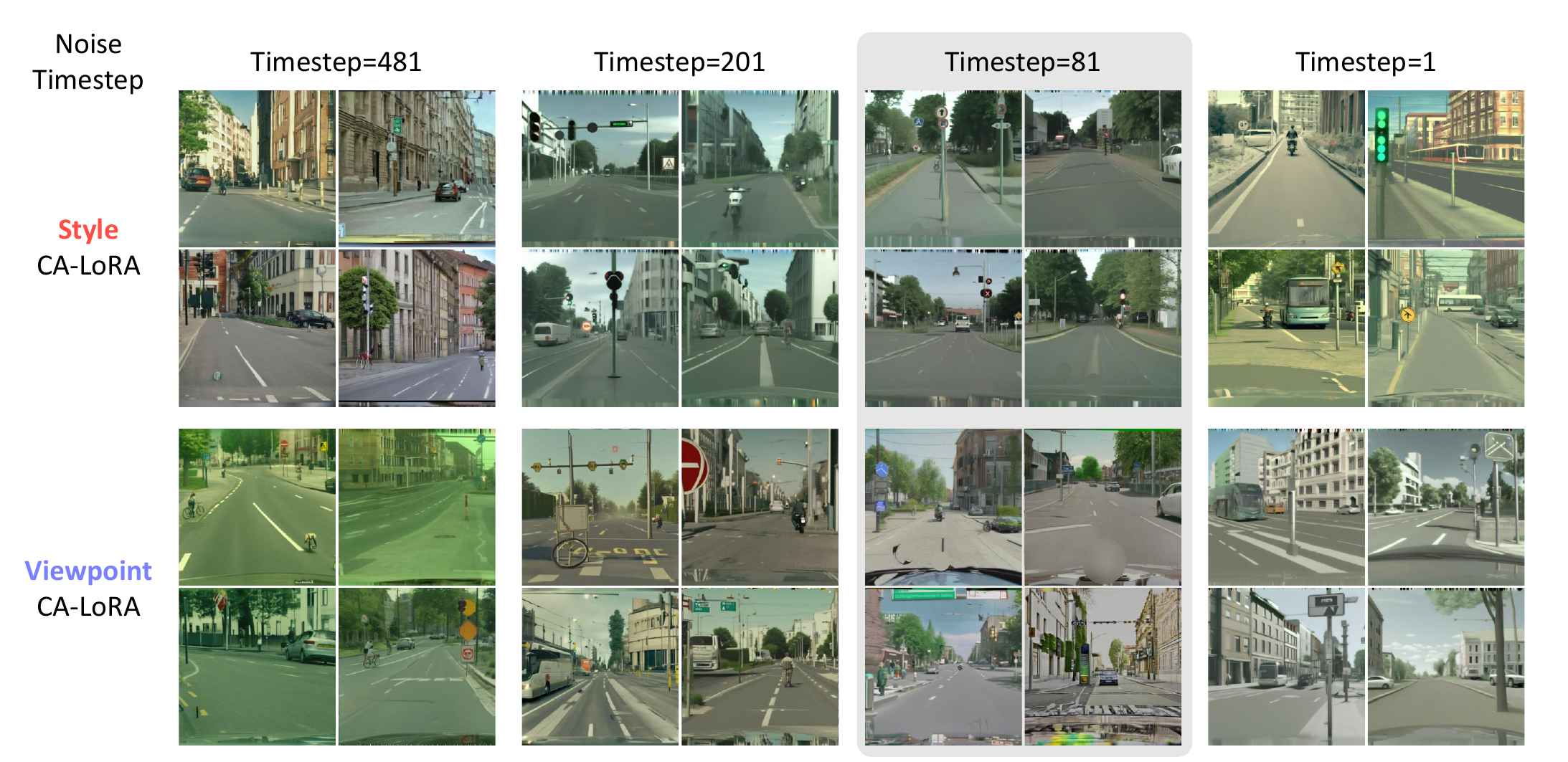}
\caption{
\textbf{Qualitative comparison across different noise timesteps.}
According to the various noise timestep, 81st timestep represents the best concept awareness, qualitatively. The style of the generated images by Style CA-LoRA is well-aligned, while the generated images by Viewpoint CA-LoRA contain diverse styles.}
\label{fig:A_concept_awareness_timestep_quali}
\end{figure*}

\subsection{Concept Awareness per Prompt Augmentation}
\label{appn:awareness_robustness}

We conduct an additional analysis of the robustness of defining desired concepts based on prompt augmentation.
As demonstrated in \Cref{sec:3_method_measure_score}, we select several hand-crafted prompt augmentations tailored to the desired concept.
While this method is flexible and can be generalized to different concepts, it may introduce instability in prompt augmentation.
To assess its robustness, we measure awareness across various prompt augmentations.
Specifically, we provide five prompt augmentations for each desired concept (style, viewpoint) derived from the original prompts, as shown below.

\begin{equation*}
\resizebox{\linewidth}{!}{$
\begin{aligned}
\textit{c}_{\text{Aug\hlred{(Style)}}} &\in \left\{ 
\parbox[c]{6.8cm}{
``\hlred{Sketch of} first-person urban street view'', \\
``\hlred{Watercolor of} first-person urban street view'', \\
``\hlred{Pop-art of} first-person urban street view'', \\
``\hlred{Line art of} first-person urban street view'', \\
``\hlred{Oil painting of} first-person urban street view''
}
\right\}
\\[3pt]
\textit{c}_{\text{Aug\hlblue{(Viewpoint)}}} &\in \left\{
\parbox[c]{6.8cm}{
``Photorealistic urban street \hlblue{in top-down view}'', \\
``Photorealistic urban street \hlblue{in high angle view}'', \\
``Photorealistic urban street \hlblue{in low angle view}'', \\
``Photorealistic urban street \hlblue{in eye-level view}'', \\
``Photorealistic urban street \hlblue{in close-up view}''
}
\right\}
\end{aligned}
$}
\end{equation*}

We then calculate the style and viewpoint awareness for each prompt augmentation, as shown in \Cref{fig:A_awareness_robustness}. As illustrated in the figure, our proposed method consistently demonstrates high awareness to similar regions across all prompt augmentations for styles. Similarly, for viewpoints, augmentations such as top-down, high-angle, and low-angle were applied, and the results indicate that our method highlights similar regions regardless of the specific viewpoint prompt. Based on these findings, we manually select the first three prompts for each desired concept. Nevertheless, developing an automated approach to search for prompt augmentations by leveraging recent LLMs could be a promising direction for enhancing concept awareness.

\subsection{Concept Awareness per Noise Timestep}
\label{appn:awareness_noise_timestep}

The amount of added noise, defined by the timestep $t$, is critical hyperparameter for calculating concept awareness.
We conduct comprehensive experiments to evaluate concept awareness in relation to the noise timestep.
Visualizations of concept awareness across different noise timesteps, along with qualitative and quantitative results, are provided.
These experimental findings offer valuable insights into the behavior of concept awareness.

\paragraph{Visualization (\Cref{fig:A_concept_awareness_timestep})}
According to our experiment, calculating concept awareness at large timesteps (noisy images) does not yield meaningful information about concept awareness.
For example, style and viewpoint sensitivities appear similar when the timestep is set to 481 out of 1000, as shown in the first column of \Cref{fig:A_concept_awareness_timestep}.
This occurs because concept-aware layers are less responsive to noisy inputs, which have a high potential to generate any image.
Conversely, extremely small timesteps (\eg, 1) also fail to capture concept awareness, as the loss from almost clean images does not provide sufficient generative information.
Therefore, we explored intermediate timesteps (\eg, 201, 81) and found that the 81st timestep reveals distinct concept sensitivities for style and viewpoint.

\paragraph{Qualitative Results (\Cref{fig:A_concept_awareness_timestep_quali})}
Additionally, we fine-tuned 2\% of the selected ratio using each concept awareness and generated images to qualitatively compare results across different noise timesteps.
As shown in \Cref{fig:A_concept_awareness_timestep_quali}, the images generated using intermediate timesteps (201, 81) better align with the intended style and viewpoint.

\paragraph{Quantitative Results (\Cref{tab:concept_awareness_cmmd})}
Finally, we quantitatively compared the intermediate timesteps using the image domain alignment metric, CMMD ($\downarrow$)~\cite{cmmd}, to evaluate the 201st and 81st timesteps.
The results indicate that the 81st timestep is the most effective for measuring concept awareness, as shown in \Cref{tab:concept_awareness_cmmd}.
While our approach selects a single timestep to measure concept awareness, averaging multiple timesteps could improve the precision and robustness of concept awareness, which may be a promising direction for future research.

\begin{table}[t]
\centering
\caption{
\textbf{Timestep ablation.}
The CMMD ($\downarrow$) and final segmentation performance of the CA-LoRA across the extracted timesteps.
}
\label{tab:concept_awareness_cmmd}

\resizebox{0.7\linewidth}{!}{
\begin{tabular}{ccccc}

\toprule
Timestep & CMMD ($\downarrow$) & Performance ($\uparrow$) \\ \midrule
1 & 2.383 & 40.49 \\
81 & \textbf{1.420} & \textbf{44.13} \\
201 & \underline{1.556} & \underline{41.68} \\
481 & 1.920 & 40.80 \\
\bottomrule

\end{tabular}
}
\end{table}

\subsection{Layer selection strategy of CA-LoRA}
\label{appn:granularity}

\begin{figure}[t]
\centering
\includegraphics[width=0.85\linewidth]{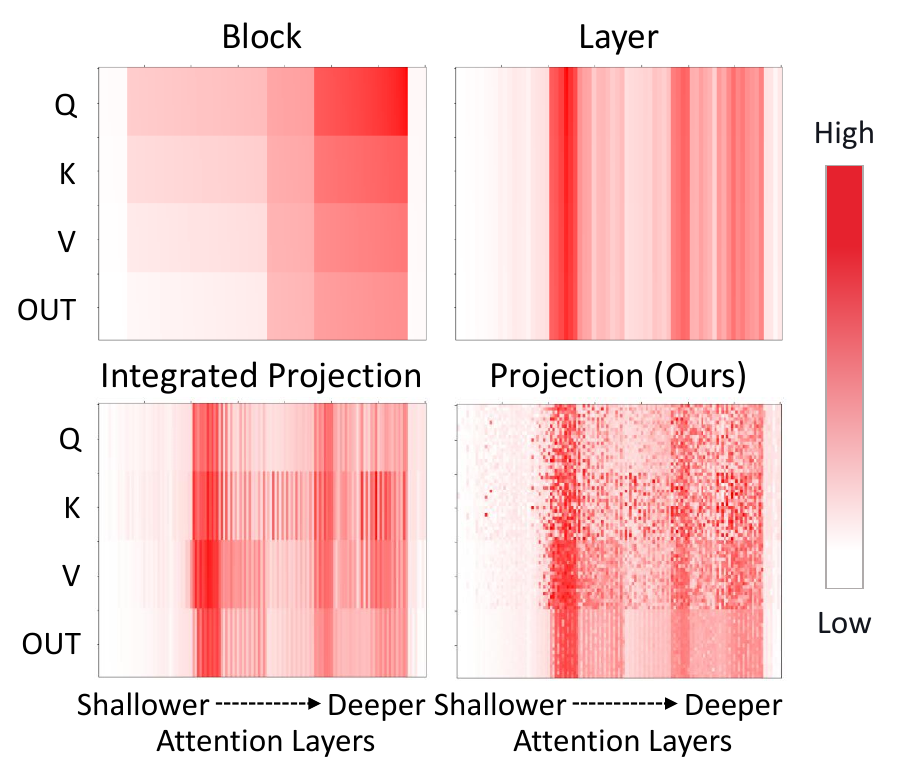}
\caption{
\textbf{Visualizing the concept awareness (style) according to various grouping granularity.}
}
\label{fig:A_granularity}
\end{figure}

\paragraph{Grouping Granularity of CA-LoRA (\cref{fig:A_granularity,tab:A_granularity})}
As illustrated in \Cref{fig:A_granularity}, we explore the performance of CA-LoRA variants that partition the base weights differently.
As shown in \cref{tab:A_granularity}, these variants generally outperform the baselines.
Among them, projection-wise CA-LoRA achieves the highest performance compared to other variants. This suggests that increasing the granularity of the base partition generally enhances performance, as it allows for more fine-grained selection of sensitive weights.

\begin{table}[t]
\centering
\caption{
\textbf{Granularity ablation.}
Number of trainable parameters and final segmentation performance by changing grouping granularity.
Only 2\% of projection-wise parameters can achieve the best results across all grouping granularity.
}
\label{tab:A_granularity}
\resizebox{\linewidth}{!}{
\begin{tabular}{lrc}
\toprule
Grouping Granularity & \% Params. & Performance ($\uparrow$) \\
\drule
No Finetuning (DatasetDM) & 0\% & 42.82 \\ \midrule
Block-wise & 32\% & \underline{43.67} {\small\color{red}(+0.85)} \\
Layer-wise & 4.8\% & {43.26 {\small\color{red}(+0.44)}} \\
Integrated Projection-wise & 5\% & {43.63 {\small\color{red}(+0.81)}} \\
Projection-wise (Ours) & 2\% & \textbf{44.13 {\small\color{red}(+1.31)}} \\
\bottomrule
\end{tabular}
}
\end{table}


\begin{figure*}[t]
\centering
\includegraphics[width=\linewidth]{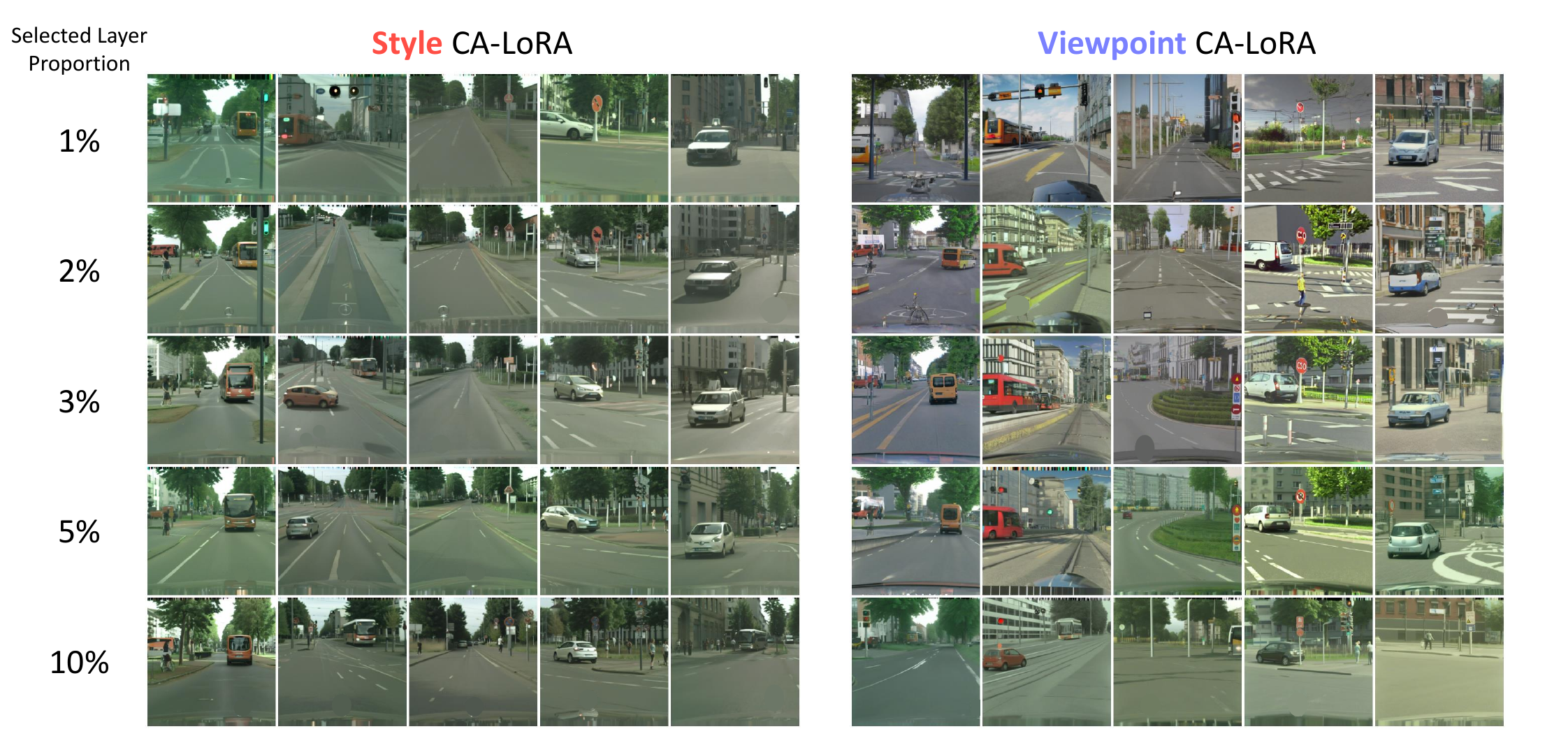}
\vspace{-10pt}
\caption{
\textbf{Qualitative results of Style- and Viewpoint CA-LoRA across different layer proportions (1\%, 2\%, 3\%, 5\%, and 10\%).}
Selecting only 2\% of the layers is already sufficient to learn the target concepts, while larger layer proportions increasingly introduce unintended concept leakage.
}
\label{fig:A_quali_layer_proportion}
\vspace{-5pt}
\end{figure*}

\vspace{10pt}

\paragraph{Layer Proportions of CA-LoRA (\Cref{fig:A_quali_layer_proportion})}
We analyze the qualitative behavior of Style CA-LoRA and Viewpoint CA-LoRA when varying the proportion of selected layers (1\%, 2\%, 3\%, 5\%, and 10\%).
The results show that selecting only 2\% of the layers is sufficient to learn the intended concepts: Style CA-LoRA produces images that accurately reflect the target style, whereas Viewpoint CA-LoRA generates images aligned with the desired viewpoint.
However, as the layer proportion increases, the model gradually begins to encode unintended, non-target concepts, indicating growing interference between concepts at higher proportions.

\section{CA-LoRA for Domain Generalization}
\label{appn:analysis_ca_lora}

In this section, we comprehensively compare the performance of CA-LoRA with baseline methods used for domain-generalized urban-scene segmentation.
We apply both image-domain alignment and image-label alignment when constructing the domain-generalization dataset, following the procedure described in \Cref{sec:4_exp_different_concept}, and also provide the comparison with DGInStyle~\cite{jia2023dginstyle}.


\begin{table}[t!]
\centering
\caption{
\textbf{Image-domain alignment in domain generalization.}
{\color{gray}$^\dagger$~DATUM utilizes one provided target-domain image per condition, whereas the other methods do not.}
}
\label{tab:A_cmmd_on_acdc}
\vspace{-0.15cm}
\resizebox{\linewidth}{!}{
\begin{tabular}{@{\hspace{0.1cm}}M{2.5cm}|M{1.4cm}M{1.6cm}M{1.4cm}M{1.4cm}|M{1.4cm}@{\hspace{0.1cm}}}
\toprule
Method & Foggy & Night-time & Rainy & Snowy & Average \\
\drule
{\color{gray} DATUM$^\dagger$} & \color{gray} \textbf{2.41} & \color{gray} \textbf{2.46} & \color{gray} \underline{2.91} & \color{gray} \textbf{2.10} & \color{gray} \textbf{2.47} \\
InstructPix2Pix & 3.43 & 3.13 & 2.99 & 3.32 & 3.22 \\
DatasetDM & 4.90 & 5.52 & 5.34 & 4.96 & 5.18 \\
Ours & \underline{2.43} & \underline{2.55} & \textbf{2.62} & \underline{2.63} & \underline{2.56} \\
\bottomrule
\end{tabular}}
\end{table}

\begin{table}[t!]
\centering
\caption{
\textbf{Image-label alignment in domain generalization.}
The first two rows report the segmentation performance of the pretrained and finetuned Mask2Former (M2F)~\cite{mask2former} on the ACDC adverse-weather dataset~\cite{acdc}. The following four rows present a comparison of image-label alignment across baseline methods, evaluated using the finetuned M2F model.
}
\label{tab:A_image_label_alignment_on_acdc_metric}
\vspace{-0.15cm}
\resizebox{\linewidth}{!}{
\begin{tabular}{@{\hspace{0.1cm}}M{2.5cm}|M{1.4cm}M{1.6cm}M{1.4cm}M{1.4cm}|M{1.4cm}@{\hspace{0.1cm}}}
\toprule
Method & Foggy & Night-time & Rainy & Snowy & Average \\
\drule
Pretrained M2F & 67.66 & 23.17 & 51.94 & 47.55 & 47.58 \\
Finetuned M2F & \textbf{78.54} & \textbf{52.16} & \textbf{66.23} & \textbf{74.79} & \textbf{67.93} \\
\drule
\color{gray} DATUM & \color{gray} - & \color{gray} - & \color{gray} - & \color{gray} - & \color{gray} - \\
InstructPix2Pix & 25.98 & \textbf{48.60} & \textbf{63.04} & \underline{40.66} & \textbf{44.57} \\
DatasetDM & \underline{40.84} & 35.90 & 47.43 & \textbf{44.02} & 42.05 \\
Ours & \textbf{41.55} & \underline{43.07} & \underline{48.69} & 39.47 & \underline{43.20} \\
\bottomrule
\end{tabular}}
\end{table}

\paragraph{Image Domain Alignment (\Cref{tab:A_cmmd_on_acdc})}
For domain generalization in urban-scene segmentation, we generated urban-scene images under various adverse weather conditions (\eg, ``foggy'', ``night-time'', ``rainy'', and ``snowy'').  
In this section, we assess the domain gap between our generated adverse weather conditions and the real ACDC~\cite{acdc} dataset.
Quantitatively, we used CMMD~\cite{cmmd} to measure image domain alignment, and the results are presented in \Cref{tab:A_cmmd_on_acdc}.
These results show that the proposed generated dataset demonstrates a significant performance gap over DatasetDM and InstructPix2Pix.
More importantly, it achieves competitive performance with DATUM, which requires training \textit{individual models} separately for each weather condition \textit{using a target domain image from the ACDC dataset}.

\paragraph{Image-Label Alignment 
(\Cref{tab:A_image_label_alignment_on_acdc_metric})}

We compare image-label alignment to assess how reliably label maps can be produced for domain-generalization datasets.
Because the generated images do not have real ground-truth annotations, we instead use pseudo ground-truth produced by a highly accurate segmentation model. As no off-the-shelf urban-scene segmentation model achieves consistently strong performance across diverse domains, we manually finetune a reliable pretrained segmentor (Mask2Former~\cite{mask2former}) on the ACDC dataset~\cite{acdc}. The first two rows in the figure show the segmentation performance on the ACDC validation set before and after finetuning, respectively.

The results of measuring image-label alignment using the fine-tuned M2F models are provided in \Cref{tab:A_image_label_alignment_on_acdc_metric}.
As shown in the table, our approach achieves superior image-label alignment compared to DatasetDM also on domain generalization setting.
InstructPix2Pix, which directly reuses Cityscapes labels and applies only minor weather edits in image, exhibits an advantage in image-label alignment.
However, despite this strong alignment, we previously highlighted its limited performance gains in \Cref{tab:exp_cityscapes,tab:exp_dg}, which we attribute to the lack of scene diversity inherent to its use of fixed segmentation label maps (see \Cref{fig:4_exp_quali_comparison}).

\begin{figure*}[t]
\centering
\includegraphics[width=0.95\linewidth]{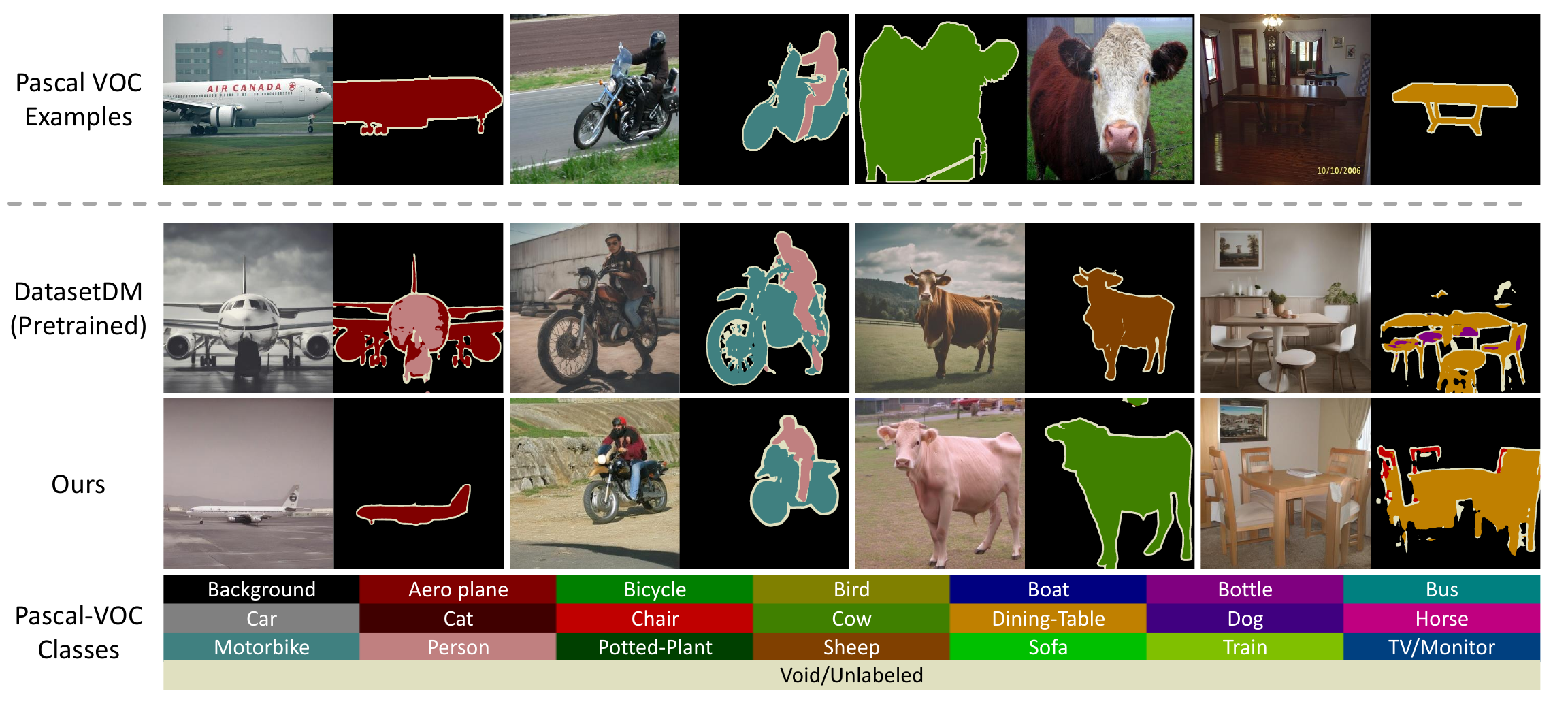}
\caption{
\textbf{Qualitative comparison for generating Pascal-VOC dataset.} Although both DatasetDM and ours are trained on the 100 labeled samples, our generated dataset shows better image domain alignment with the original Pascal-VOC examples and also shows better image-label alignment.
}
\label{fig:A_voc_quali}
\end{figure*}

\begin{table}[t]
\centering
\caption{
\textbf{Comparison with DGInStyle of generated datasets for domain generalization of urban-scene segmentation (mIoU)}.
We highlight the performance improvements are larger than DGInStyle, especially on HRDA backbone.
{\small \color{gray} *The gray color indicates the reported score from the DGInStyle authors.}
}
\label{tab:A_comparison_dginstyle}
\vspace{-5pt}
\resizebox{\linewidth}{!}{
\begin{tabular}{@{\hspace{0.1cm}}M{2.1cm}|M{2.1cm}|M{1.3cm}M{1.1cm}M{1.1cm}M{1.1cm}|M{2.1cm}@{\hspace{0.1cm}}}

\toprule
DG Method & Method & ACDC & DZ & BDD & MV & Average \\
\drule

\multirow{4}{*}[-0.1cm]{ColorAug} 
 & {\color{gray} Baseline} & {\color{gray} 52.38} & {\color{gray} 23.00} & {\color{gray} 53.33} & {\color{gray} 60.06} & {\color{gray} 47.19} \\
 & {\color{gray} DGInStyle} & {\color{gray} 55.19} & {\color{gray} 26.83} & {\color{gray} 55.18} & {\color{gray} 59.95} & {\color{gray} 49.29 {\small{(+2.10)}}} \\ \cmidrule(lr){2-7}
 & Baseline & 53.12 & 25.69 & 53.00 & 59.81 & 47.91 \\
 & CA-LoRA & 56.07 & 29.75 & 54.35 & 61.40 & \textbf{50.39 {\small{\color{red}(+2.49)}}} \\ \midrule

\multirow{4}{*}[-0.1cm]{DAFormer} 
 & {\color{gray} Baseline} & {\color{gray} 55.15} & {\color{gray} 28.28} & {\color{gray} 54.19} & {\color{gray} 61.67} & {\color{gray} 49.82} \\
 & {\color{gray} DGInStyle} & {\color{gray} 57.74} & {\color{gray} 28.55} & {\color{gray} 56.26} & {\color{gray} 62.67} & {\color{gray} 51.31 {\small{(+1.48)}}} \\ \cmidrule(lr){2-7}
 & Baseline & 53.98 & 27.82 & 54.29 & 62.69 & 49.70 \\
 & CA-LoRA & 55.83 & 31.68 & 54.68 & 63.09 & \textbf{51.32 {\small{\color{red}(+1.63)}}} \\ \midrule

\multirow{4}{*}[-0.1cm]{HRDA} 
 & {\color{gray} Baseline} & {\color{gray} 59.70} & {\color{gray} 31.07} & {\color{gray} 58.49} & {\color{gray} 68.32} & {\color{gray} 54.40} \\
 & {\color{gray} DGInStyle} & {\color{gray} 61.00} & {\color{gray} 32.60} & {\color{gray} 58.84} & {\color{gray} 67.99} & {\color{gray} 55.11 {\small{(+0.71)}}} \\ \cmidrule(lr){2-7}
 & Baseline & 58.48 & 29.46 & 56.12 & 64.27 & 52.08 \\
 & CA-LoRA & 58.93 & 34.41 & 56.56 & 64.54 & \textbf{53.61 {\small{\color{red}(+1.53)}}} \\
\bottomrule

\end{tabular}
} 
\end{table}

\paragraph{Comparison with DGInStyle (\Cref{tab:A_comparison_dginstyle})}

As shown in \Cref{tab:A_comparison_dginstyle}, our CA-LoRA achieves performance gains comparable to, and in most cases larger than, those reported for DGInStyle~\cite{jia2023dginstyle}.
As their code and Cityscapes-generated data are unavailable, and their baseline accuracy is marginally higher than ours, we compare only the performance gains rather than the absolute numbers.
We will update the comparison once the official code becomes available.
Notably, our approach consistently exhibits substantial improvements when applied to the HRDA method.
In contrast, DGInStyle generates images using fixed label maps, which limits its ability to construct semantically diverse scenes. We hypothesize that this restricted diversity contributes to the relatively small performance gains observed when DGInStyle is used with advanced DG methods such as HRDA.

\section{Experiments in General Domain}
\label{appn:voc}

Since our primary goal is to cover urban-scene segmentation, we focused on style and viewpoint as the desired concepts and conducted experiments exclusively on urban-scene datasets such as Cityscapes.
However, the CA-LoRA methodology is not limited to urban-scene datasets. It can also be applied to general datasets for in-domain segmentation dataset generation. 
In this section, we demonstrate experiments on the Pascal-VOC dataset~\cite{everingham2010pascal}, showcasing how our approach improves few-shot semantic segmentation performance.

\subsection{Experimental setup}
In this experiment, we trained on a total of 100 real image-label pairs and evaluated the model using the 1,449 images in the Pascal-VOC validation set. For the text-to-image generation model, we applied the same style awareness score used in the Cityscapes experiment, setting the selected proportion to 10\%. 
During the training of the text-to-image generation model, the prompt "a photo" was used. For training the label generator and generating the dataset, the prompt "a photo of a \{class names\}" was employed. The label generator was trained with a batch size of 4 for 90K iterations, ultimately producing 2,000 image-label pairs.
When utilizing the generated dataset, the process was consistent with the in-domain semantic segmentation experiments.
Specifically, Mask2Former was trained on the real dataset for 90K iterations (Baseline), followed by fine-tuning on the combined real and generated dataset for an additional 30K iterations. 
All other hyper-parameters remained identical to those used in the Cityscapes in-domain semantic segmentation experiment, as detailed in \Cref{tab:A_hparam1,tab:A_hparam2,tab:A_hparam3}.
Notably, the diffusion timestep ($t=81$) and the real-to-synthetic data ratio ($1:1$) are directly transferred from the Cityscapes experiment without re-tuning, demonstrating the robustness of our key hyperparameters across datasets.

\subsection{Quantitative and Qualitative Results}
As shown in \Cref{tab:A_exp_voc}, using Style CA-LoRA on the Pascal-VOC dataset resulted in a performance improvement of 0.93 mIoU. In contrast, DatasetDM, which omitted the fine-tuning process for the text-to-image generation model, showed a performance drop of 8.43 mIoU. This highlights the importance of concept-aware finetuning for style, even in general datasets beyond urban-scene datasets.
\Cref{fig:A_voc_quali} provides further insight into the role of style information. A significant image domain gap is evident between the images generated by the pretrained text-to-image generation model and the dataset generated using Pascal-VOC. This demonstrates the impact of image domain alignment. Quantitatively, the CMMD, which was 1.46 for the pretrained model, decreased to 0.81 after alignment, illustrating the reduced domain gap and its contribution to performance improvement.

\begin{figure}[t]
\centering
\includegraphics[width=\linewidth]{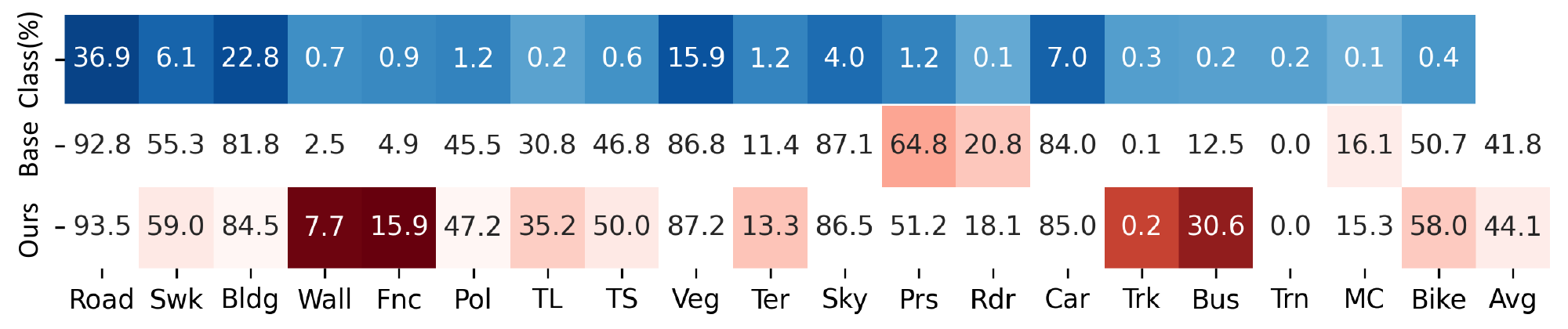}
\includegraphics[width=\linewidth]{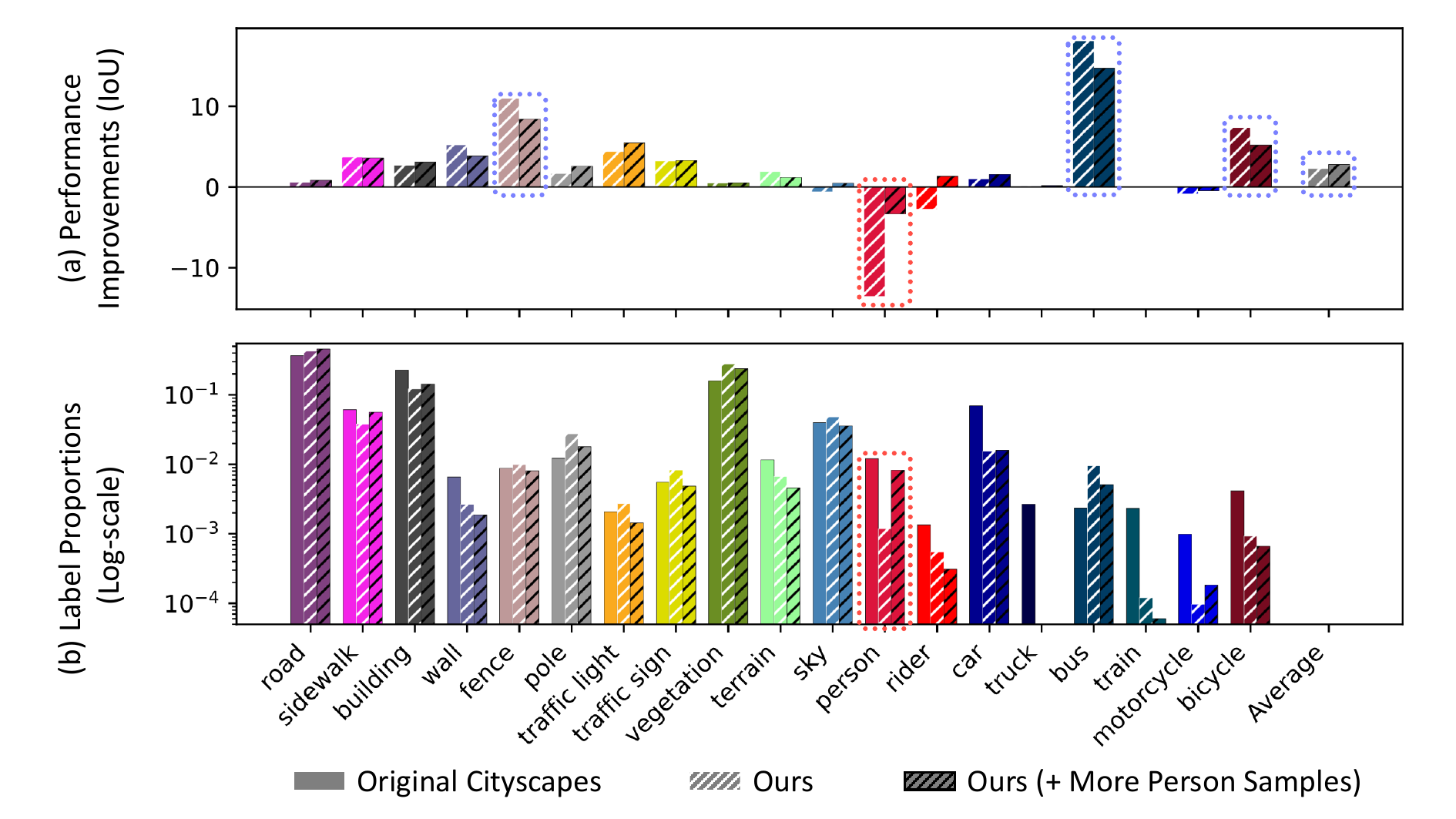}
\caption{
\textbf{Class-wise performance improvements and label proportions.}
(a) Class-wise IoU improvements and (b) label proportions for Cityscapes, Ours, and Ours (+ More Person Samples).
Adding 500 additional “person" samples balances label proportions. Rare classes such as “bus", “fence", and “bicycle" show clear gains, while decreases in “person" and “rider" arise from insufficient samples. Supplementing these classes further leads to more balanced improvements and a higher overall average.
}
\label{fig:A_classwise_iou}
\end{figure}

\begin{table}[t!]
\centering
\caption{
\textbf{In-domain segmentation performance (mIoU) of the Pascal VOC dataset.}
In the first row, we report the performance of Mask2Former trained on different fractions of the PASCAL VOC dataset (Baseline). While DatasetDM often degrades segmentation performance due to incorrect labels, CA-LoRA consistently improves the results.
}
\label{tab:A_exp_voc}
\vspace{-5pt}
\resizebox{\linewidth}{!}{
\begin{tabular}{@{\hspace{0.1cm}}M{2.8cm}|M{3.5cm}M{3.5cm}@{\hspace{0.1cm}}}
\toprule
\multirow{2}{*}[-0.1cm]{Method} & \multicolumn{2}{c}{Fraction of the PASCAL VOC Dataset} \\ \cmidrule(lr){2-3}
 & 100 images ($\sim$7\%) & 1,464 images (100\%) \\
\drule
Baseline & 44.39 & 72.54 \\ \midrule
DatasetDM & 36.16 \small{\color{blue} (-8.43)} & 70.78 \small{\color{blue} (-1.76)} \\
CA-LoRA (Ours) & \textbf{45.52 \small{\color{red} (+0.93)}} & \textbf{73.72 \small{\color{red} (+1.18)}} \\
\bottomrule
\end{tabular}
}
\end{table}

\section{Class-Specific Dataset Generation}
\label{appn:class_specific_generation}

\begin{figure*}[t]
\centering
\includegraphics[width=\linewidth]{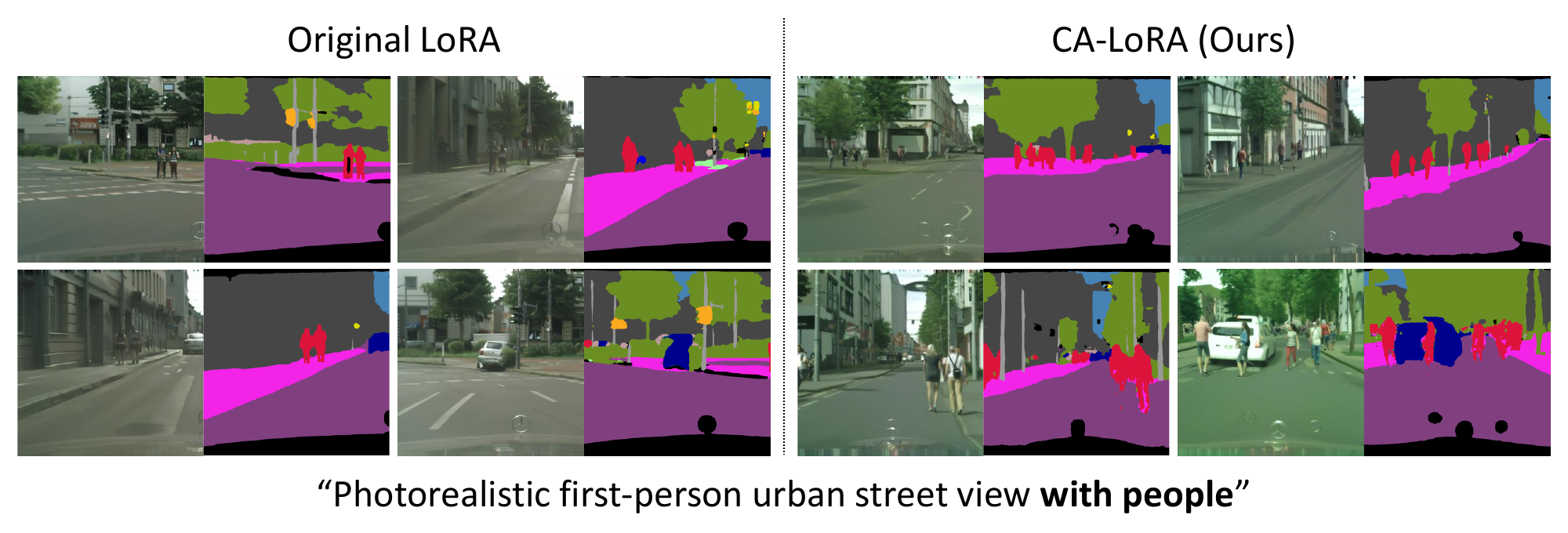}
\vspace{-0.7cm}
\caption{
\textbf{Qualitative comparison between the original LoRA and CA-LoRA for generating “person" samples.}
While the original LoRA produces limited variations resembling training images, CA-LoRA generates more diverse “person" scenes by learning only the style from the source dataset.
}
\label{fig:A_classwise_iou_quali}
\end{figure*}

\begin{figure*}[t]
\centering
\includegraphics[width=\linewidth]{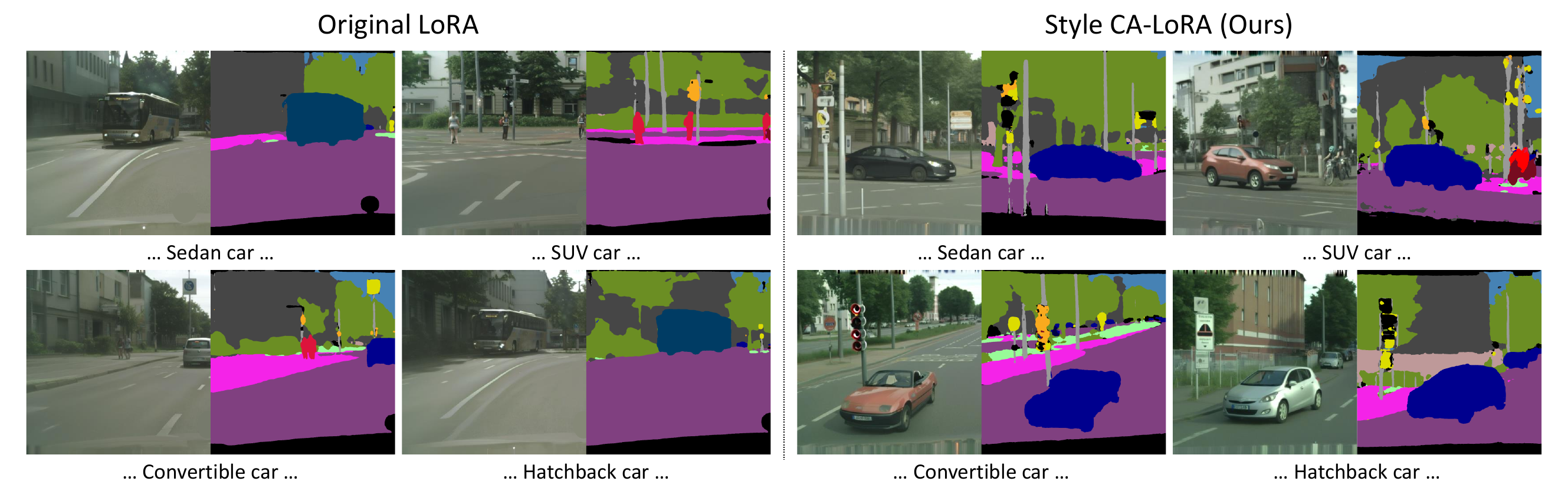}
\vspace{-0.7cm}
\caption{
Generated image-label pairs showcasing various styles of cars, including sedan, SUV, convertible, and hatchback.
Unlike the Original LoRA, since the Style CA-LoRA exclusively learned only the style from the Cityscapes, we can generate various types of cars in Cityscapes-style.
}
\label{fig:A_various_cars}
\end{figure*}

\begin{figure*}[t]
\centering
\includegraphics[width=\linewidth]{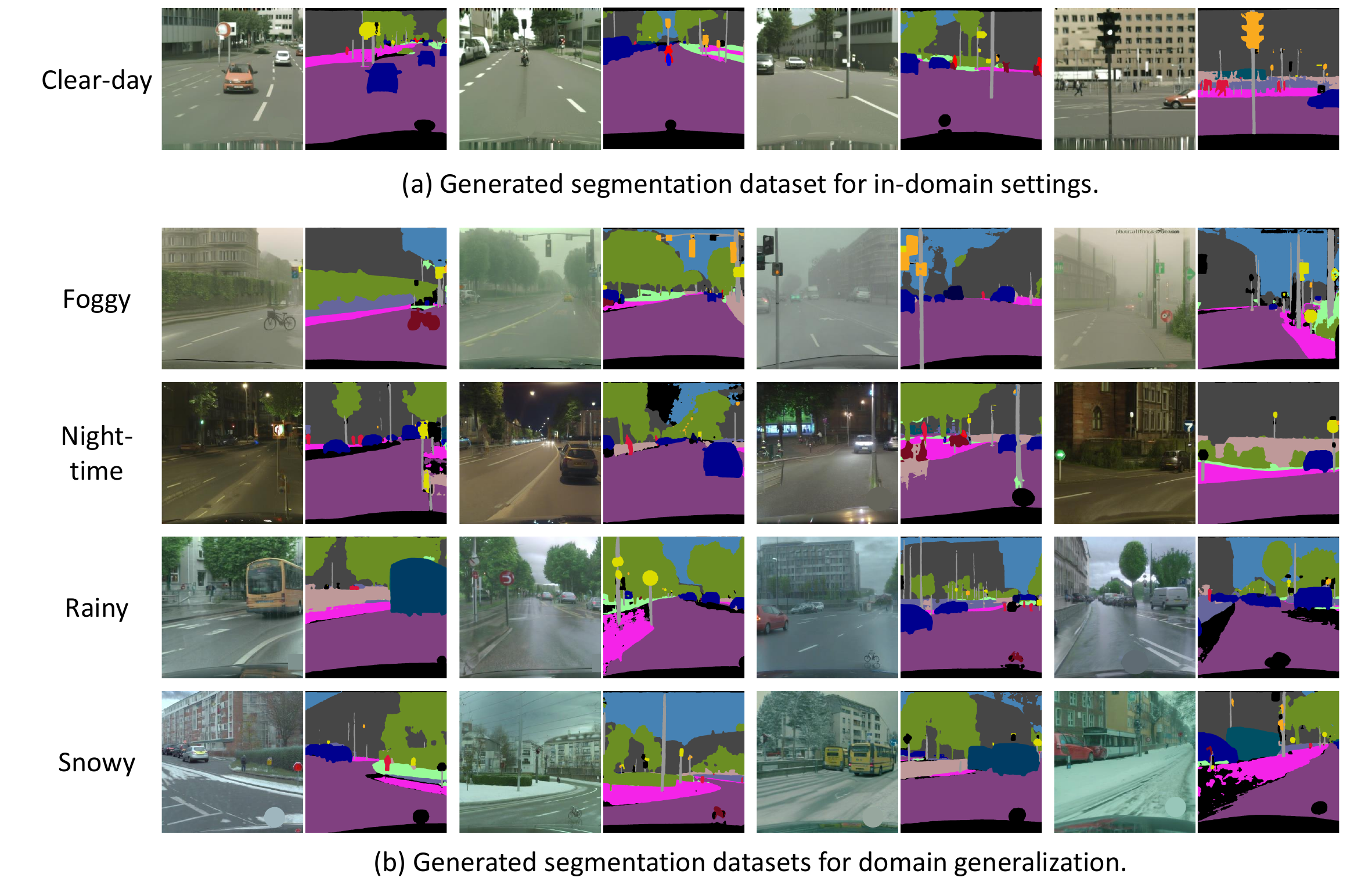}
\vspace{-0.7cm}
\caption{
Additional qualitative results
}
\label{fig:A_additional_quali}
\end{figure*}

\subsection{Class-wise Segmentation Performance}

In this section, we present a detailed analysis of class-wise improvements, highlighting the effectiveness of the proposed method, particularly for rare classes. Additionally, we introduce a class-balanced performance improvement strategy tailored to specific classes.

\paragraph{Class-wise Performance Improvements (\Cref{fig:A_classwise_iou})}
Urban-scene segmentation has distinct challenges, including class imbalance and co-occurrence issues~\cite{kim2024weakly}, making class-wise analysis particularly important. 
We present the class-wise IoU improvements in \Cref{fig:A_classwise_iou}. 
As shown in \Cref{fig:A_classwise_iou} (a), our proposed dataset generation approach proves especially effective for rare classes such as ``bus'', ``fence'', and ``bicycle''.
However, the generated dataset often fails to improve performance in certain classes, such as ``person'' and ``rider''. 
As illustrated in \Cref{fig:A_classwise_iou} (b), this degradation is primarily due to the insufficient number of generated samples for the ``person'' class.
Since the synthetic dataset is generated randomly, disparities in label proportions can occur.
To mitigate this, we propose a simple yet effective technique to increase the proportion of the target class.

\begin{table}[t]
\centering
\caption{
\textbf{In-domain segmentation performance of datasets incorporating the diverse cars dataset.} Incorporating the diverse cars dataset especially improved performance for vehicle classes such as ``car'', ``bus'', and ``motorcycle'', leading to overall performance improvements.
}
\label{tab:A_various_cars}
\vspace{-5pt}
\resizebox{\linewidth}{!}{
\begin{tabular}{@{\hspace{0.1cm}}L{3.3cm}|M{1.3cm}M{1.3cm}M{1.6cm}|M{1.5cm}@{\hspace{0.1cm}}}
\toprule
\multirow{2}{*}[-0.1cm]{Method} & \multicolumn{3}{c|}{IoU} & \multirow{2}{*}[-0.1cm]{mIoU} \\ \cmidrule(lr){2-4}
 & Car & Bus & Motorcycle &  \\
\drule
Baseline & 84.02 & 12.51 & \underline{16.11} & 41.83 \\ \midrule
Ours & \underline{85.03} & \underline{30.59} & 15.26 & \underline{44.12} \\
Ours (+ Diverse Cars) & \textbf{85.34} & \textbf{32.48} & \textbf{17.77} & \textbf{44.95} \\
\bottomrule

\end{tabular}}
\vspace{-5pt}
\end{table}

\paragraph{Segmentation Dataset Generation Focused on a Specific Class (\Cref{fig:A_classwise_iou,fig:A_classwise_iou_quali})}
As detailed in \Cref{sec:3_method_generation} and \Cref{tab:A_hparam1}, we generated the dataset using the prompt ``photorealistic first-person urban street view with [Class names]'', where the class names were extracted from the label map of the training set by retrieving the names of all classes present in the label map. 
While the synthesized text prompt partially reflects the label proportions of the training set, it does not strictly enforce these proportions.
As a result, the proposed generated dataset may exhibit misaligned label proportions, as illustrated in \Cref{fig:A_classwise_iou} (b).

To address this issue, we propose a class-specific generation approach that manually increases the target class by modifying the generation prompts.
Specifically, we generated an additional 500 samples using the prompt ``photorealistic first-person urban street view with people'' to increase the proportion of the ``person'' class.\footnote{We also experimented with ``photorealistic first-person urban street view with person'', but using ``people'' as the test prompt proved to be more effective in increasing the label proportion for the ``person'' class.}
Since we selectively fine-tuned the LoRA to learn only the style from Cityscapes, it enables effective manipulation using the text prompt, which the original LoRA cannot achieve, as demonstrated in \Cref{fig:A_classwise_iou_quali}.
As illustrated in \Cref{fig:A_classwise_iou} (b), this approach successfully increased the proportion of the ``person'' class and mitigated its performance degradation. Furthermore, as shown in \Cref{fig:A_classwise_iou} (a), this adjustment led to additional performance improvements, increasing the average IoU from 44.12 to 44.59.

\subsection{Generating Datasets with Diverse Class Names}

Since the Style CA-LoRA selectively fine-tuned only the style from the in-domain Cityscapes dataset, it retains its generalization ability for text prompts such as objects.
Leveraging this capability, we aim to generate a broader variety of images using more diverse class names beyond those provided in the dataset.
In this experiment, we refined the prompts for generating images previously created with the simple class name ``car'' by subdividing them into ``sedan car'', ``SUV car'', ``convertible car'', and ``hatchback car'', as shown in \Cref{fig:A_various_cars}.
As illustrated in the figure, while the original LoRA fine-tuned text-to-image generation model struggles to produce diverse styles of cars, our approach reliably generates a wide variety of cars that align with the test prompts.

We then conducted an in-domain few-shot experiment (Cityscapes 0.3\%) using the additional diverse cars dataset, following the experimental setup described in~\Cref{sec:4_exp_setup}.
As shown in \Cref{tab:A_various_cars}, incorporating the diverse cars dataset significantly improves segmentation performance, particularly for vehicle classes.
Beyond generating diverse cars, applying textual augmentations to other class names for dataset creation represents a promising direction for advancing segmentation dataset generation.

\clearpage

\begin{table*}[t]
\centering
\vspace{1cm}
\caption{
Hyperparameters to fine-tune Stable Diffusion XL~\cite{sdxl}. 
The class names are extracted from the label map in the training set by retrieving the names of all classes that appear in the label map.
}
\label{tab:A_hparam1}
\begin{tabular}{l|r}

\toprule
Hyperparameter & Value \\ \drule
Rank & 64 \\
Learning rate & 1e-4 \\
Batch size & 1 \\
Training iteration & 10K \\
Data augmentation & Random horizontal flip, Random crop \\
Resolution & (1024, 1024) \\
Learning rate scheduler & constant \\
Optimizer & AdamW~\cite{adamw} \\
Adam beta1 & 0.9 \\
Adam beta2 & 0.999 \\
Adam weight decay & 0.01 \\
Training prompt & ``photorealistic first-person urban street view'' \\
\midrule \multicolumn{2}{c}{Test-time hyperparameters} \\ \midrule
Num. inference steps & 25 \\
Guidance scale & 5.0 \\
Test prompt augmentation (In-domain) & ``... with [Class names]''\protect\footnotemark \\
Test prompt augmentation (DG) & ``... in [Weather Condition] with [Class names]'' \\
\bottomrule

\end{tabular}
\end{table*}

\begin{table*}[t]
\centering
\caption{Hyperparameters to train label generator followed by DatasetDM~\cite{datasetdm}. 
}
\label{tab:A_hparam2}
\resizebox{0.8\textwidth}{!}{
\begin{tabular}{l|r}

\toprule
Hyperparameter & Value \\ \drule
Architecture & Mask2Former-shaped label generator~\cite{datasetdm} \\
Learning rate & 1e-4 \\
Batch size & 2 for all few-shot, 8 for fully-supervised \\
Training iteration (few-shot) & 12k, 24k, 24k, and 48k for 0.3\%, 1\%, 3\%, and 10\%, respectively \\
Training iteration (fully-supervised) &  90K \\
Data augmentation & Random horizontal flip, Random resized crop (0.5, 2.0) \\
Resolution & (1024, 1024) \\
Learning rate scheduler & PolynomialLR(power=0.9) \\
Optimizer & Adam~\cite{kingma2014adam} \\
Adam beta1 & 0.9 \\
Adam beta2 & 0.999 \\
Adam weight decay & 0.0 \\
\bottomrule

\end{tabular}
}
\end{table*}

\begin{table*}[t]
\centering
\caption{Hyperparameters to fine-tune Mask2Former~\cite{mask2former}. We modify the learning rate, batch size and training iteration from the original Mask2Former training configuration.}
\label{tab:A_hparam3}
\begin{tabular}{l|r}

\toprule
Hyperparameter & Value \\ \drule
Model Architecture & Mask2Former~\cite{mask2former} \\
Num. generated images & 500 for all few-shot, and 3,000 for fully-supervised \\
Learning rate & 3e-6 \\
Batch size & 2 for all few-shot, and 8 for fully-supervised \\
Mixed batch & real:syn = 1:1 \\
Training iteration & 30K \\
Data augmentation & Random horizontal flip, Random resized crop (0.5, 2.0) \\
Resolution & (512, 1024) \\
Learning rate scheduler & PolynomialLR(power=0.9) \\
Optimizer & AdamW~\cite{adamw} \\
Adam beta1 & 0.9 \\
Adam beta2 & 0.999 \\
Adam weight decay & 0.05 \\
\bottomrule

\end{tabular}
\end{table*}

\begin{table*}[t]
\centering
\caption{Hyperparameters to train domain generalization in segmentation including ColorAug, DAFormer~\cite{hoyer2022daformer}, and HRDA~\cite{hoyer2022hrda}, followed by DGInStyle~\cite{jia2023dginstyle}.}
\label{tab:A_hparam4}
\resizebox{0.9\textwidth}{!}{
\begin{tabular}{l|ccc}

\toprule
Hyperparameter & Value (ColorAug)~\cite{xie2021segformer} & Value (DAFormer)~\cite{hoyer2022daformer} & Value (HRDA)~\cite{hoyer2022hrda} \\ \drule
Model Architecture & SegFormer & DAFormer & HRDA \\
Backbone &  & MiT-B5~\cite{xie2021segformer} &  \\
Num. generated images & \multicolumn{3}{c}{500 for each weather condition (clear, foggy, night-time, rainy, and snowy)} \\
Learning rate &  & 6e-5 &  \\
Batch size &  & 2 &  \\
Training iteration &  & 40K &  \\
Data augmentation for Gen. & \multicolumn{3}{c}{Random horizontal flip, PhotoMetricDistortion} \\
Data augmentation for Real & \multicolumn{3}{c}{Random horizontal flip, Random crop, DACS~\cite{tranheden2021dacs}} \\
Resolution & (512, 512) & (512, 512) & (1024, 1024) \\
Learning rate scheduler &  & PolynomialLR(power=0.9) &  \\
Learning rate warmup &  & Linear &  \\
Learning rate warmup iteration &  & 1500 &  \\
Learning rate warmup ratio &  & 1e-6 &  \\
Optimizer &  & AdamW~\cite{adamw} &  \\
Adam beta1 &  & 0.9 &  \\
Adam beta2 &  & 0.999 &  \\
Adam weight decay &  & 0.01 &  \\
SHADE & False & True & True \\
RCS~\cite{hoyer2022daformer, hoyer2022hrda} &  & False &  \\
\bottomrule

\end{tabular}
}
\end{table*}

\begin{algorithm*}[t]
\caption{PyTorch-like Pseudocode of Concept Awareness}
\label{algo:concept_awareness}
\PyComment{pipe: text-to-image generation diffusers pipeline} \\
\PyComment{c: str = "photorealistic first-person urban street view"} \\
\PyComment{c\_augs: List[str] = List of the augmented prompts}\\
\PyComment{t: int = pre-defined timestep} \\
\PyComment{n\_img: int = number of generated images for average} \\

\PyCode{unet = pipe.unet}\\
\PyCode{unet = unet.requires\_grad\_(True)}\\

\PyComment{optimizer for clear gradients}\\
\PyCode{optimizer = torch.optim.AdamW(list(filter(lambda p: p.requires\_grad, unet.parameters())))}\\

\PyCode{imgs = [pipe(c).images[0] \PyDef{for} \_ \PyDef{in} n\_img]  \PyComment{generate images}} \\

\PyCode{awareness = []}\\

\PyDef{for} \PyCode{img} \PyDef{in} \PyCode{imgs:}    
\PyComment{average over generated images}

\Indp
\PyDef{for} \PyCode{c\_aug} \PyDef{in} \PyCode{c\_augs:}    \PyComment{average over augmented captions}

\Indp
\PyCode{latent = pipe.vae.encode(img)}\\
\PyCode{noise = torch.randn\_like(latent)}\\
\PyCode{noisy\_latent = pipe.scheduler.add\_noise(latent, noise, t)}\\

\PyCode{prompt\_embeds = encode\_prompt(c)}\\
\PyCode{model\_pred = unet(noisy\_latent, t, prompt\_embeds)}\\

\PyCode{gt\_diff = noise}\\

\PyDef{with} \PyCode{torch.no\_grad():}

\Indp
\PyCode{prompt\_embeds\_aug = encode\_prompt(c\_aug)} \\
\PyCode{gt\_concept = unet(noisy\_latent, t, prompt\_embeds\_aug)}\\

\Indm
\PyCode{loss\_diff = torch.nn.functional.mse\_loss(model\_pred, gt\_diff)}\\
\PyCode{loss\_concept = torch.nn.functional.mse\_loss(model\_pred, gt\_concept)}\\

\PyCode{loss\_diff.backward(retain\_graph=True)}\\
\PyCode{grads\_diff = get\_unet\_grads(unet)} \PyComment{\Cref{algo:concept_helper}}\\
\PyCode{optimizer.zero\_grad()}\\

\PyCode{loss\_concept.backward()}\\
\PyCode{grads\_concept = get\_unet\_grads(unet)} \PyComment{\Cref{algo:concept_helper}}\\
\PyCode{optimizer.zero\_grad()}\\

\PyCode{awareness.append(grads\_concept / grads\_diff)}

\Indm
\Indm

\PyCode{awareness\_avg = average\_gradients(awareness)}  \PyComment{\Cref{algo:concept_helper}}
\\

\end{algorithm*}

\begin{algorithm*}[t]
\caption{PyTorch-like Helper Functions for Concept Awareness}
\label{algo:concept_helper}

\PyDef{def} \PyCode{getattr\_recursive(module, attrs: List[str]):}

\Indp
\PyCode{target\_module = module}

\PyDef{for} \PyCode{attr} \PyDef{in} \PyCode{attrs:}

\Indp
\PyCode{target\_module = getattr(target\_module, attr)}

\Indm
\PyDef{return} \PyCode{target\_module}

\Indm
\PyCode{}

\PyDef{def} \PyCode{get\_unet\_grads(unet):}

\Indp
\PyCode{grads = \{'to\_q': [], 'to\_k': [], 'to\_v': [], 'to\_out.0': []\}}\\
\PyDef{for} \PyCode{attn\_name} \PyDef{in} \PyCode{unet.attn\_processors.keys():}

\Indp
\PyCode{attn\_module = getattr\_recursive(unet, attn\_name.split('.')[:-1])}\\

\PyDef{for} \PyCode{proj\_name} \PyDef{in} \PyCode{grads.keys():}

\Indp
\PyCode{proj = getattr\_recursive(attn\_module, proj\_name.split('.'))}\\
\PyCode{head\_dim = 1 if proj\_name == 'to\_out.0' else 0}\\
\PyCode{grads\_chunk = torch.chunk(proj.weight.grad.cpu(), attn\_module.heads, dim=head\_dim)}\\
\PyCode{grads[proj\_name].append([(grad ** 2).mean().sqrt().item() for grad in grads\_chunk])}\\

\Indm
\Indm
\PyDef{return} \PyCode{grads}
\\

\PyCode{\PyDef{def} average\_gradients(grads):}

\Indp
\PyCode{grad\_avg = \{'to\_q': [], 'to\_k': [], 'to\_v': [], 'to\_out.0': []\}}\\

\PyCode{\PyDef{for} key \PyDef{in} grad\_avg:}

\Indp
\PyCode{\PyDef{for} grad \PyDef{in} grads:}

\Indp
\PyCode{grad\_avg[key].append(grad[key])}

\Indm
\PyCode{grad\_avg[key] = torch.mean(torch.tensor(grad\_avg[key]), dim=0)}\\

\Indm
\PyCode{\PyDef{return} grad\_avg}\\

\Indm

\end{algorithm*}

\begin{algorithm*}[t]
\caption{PyTorch-like Pseudocode of Modifying forward function of CA-LoRA}
\label{algo:ca_lora_forward}
\PyComment{F: torch.nn.functional}\\

\PyDef{def} \PyCode{modify\_to\_ca\_lora(layer, reduced\_layer)}:

\Indp
\PyComment{layer: diffusers.models.LoRACompatibleLinear}\\
\PyComment{reduced\_layer: 'A' or 'B'}\\

\PyDef{def} \PyCode{ca\_lora\_set\_lora\_layer(self: LoRACompatibleLinear):}

\Indp
\PyDef{def} \PyCode{set\_lora\_layer(lora\_layer, indices):}

\Indp
\PyCode{self.lora\_layer = lora\_layer}\\
\PyDef{if} \PyCode{indices is not None:}

\Indp
\PyCode{self.indices = indices}

\Indm

\Indm
\PyDef{return} \PyCode{set\_lora\_layer}

\Indm
\PyCode{}

\PyDef{def} \PyCode{ca\_lora\_set\_forward(self: LoRACompatibleLinear):}

\Indp
\PyDef{def} \PyCode{forward(hidden\_states: torch.Tensor, scale: float = 1.0):}

\Indp
\PyDef{if} \PyCode{self.lora\_layer is None:}

\Indp
\PyDef{return} \PyCode{F.linear(hidden\_states, self.weight, self.bias)}

\Indm

\PyDef{else}\PyCode{:}

\Indp
\PyDef{if} \PyCode{self.indices is not None:}

\Indp
\PyComment{CA-LoRA (start)}\\
\PyCode{org = F.linear(hidden\_states, self.weight, self.bias)}

\PyDef{if} \PyCode{reduced\_layer == 'B':}

\Indp
\PyCode{org[:, :, self.indices] += scale * self.lora\_layer(hidden\_states)}

\Indm
\PyDef{else}:

\Indp
\PyCode{org += scale * self.lora\_layer(hidden\_states[:, :, self.indices])}

\Indm
\PyDef{return} \PyCode{org}\\
\PyComment{CA-LoRA (end)}

\Indm
\PyDef{else}:

\Indp
\PyDef{return} F.linear(hidden\_states, self.weight, self.bias) + scale * self.lora\_layer(hidden\_states)

\Indm
\Indm
\Indm
\PyDef{return} forward

\Indm
\PyCode{}\\
\PyCode{layer.set\_lora\_layer = ca\_lora\_set\_lora\_layer(layer)}\\
\PyCode{layer.forward = ca\_lora\_set\_forward(layer)}\\

\PyDef{return} \PyCode{layer}

\end{algorithm*}

\begin{algorithm*}[t]
\caption{PyTorch-like Pseudocode of selecting projection layers for CA-LoRA}
\label{algo:ca_lora_define}

\PyDef{def} \PyCode{apply\_ca\_lora(unet, selected\_layers, rank):}

\Indp
\PyDef{for} \PyCode{attn\_processor\_name} \PyDef{in} \PyCode{unet.attn\_processors.keys():}

\Indp

\PyCode{\_selected\_layers = [selected\_layer \PyDef{for} selected\_layer \PyDef{in} selected\_layers \PyDef{if} '.'.join(attn\_processor\_name.split('.')[:-1]) \PyDef{in} selected\_layer]}\\
\PyDef{if} \PyCode{len(\_selected\_layers) == 0:}

\Indp
\PyDef{continue}
    
\Indm
\PyCode{attn\_module = getattr\_recursively(unet, attn\_processor\_name.split('.')[:-1])} \\
\PyComment{getattr\_recursively: \Cref{algo:concept_helper}}\\
\PyCode{dim\_head = attn\_module.out\_dim // attn\_module.heads}

\PyCode{\PyDef{for} layer\_type \PyDef{in} ('to\_q', 'to\_k', 'to\_v', 'to\_out.0'):}

\Indp

\PyCode{selected\_layers\_proj = [selected\_layer for selected\_layer \PyDef{in} \_selected\_layers \PyDef{if} layer\_type \PyDef{in} selected\_layer]
is\_out = layer\_type == 'to\_out.0'}\\
\PyCode{\PyDef{if} len(selected\_layers\_proj) == 0:}

\Indp
\PyDef{continue}

\Indm
\PyCode{projection\_layer = getattr\_recursively(attn\_module, layer\_type.split('.'))} \\ 
\PyComment{getattr\_recursively: \Cref{algo:concept_helper}}\\

\PyComment{Projection-wise CA-LoRA (start)}\\
\PyCode{head\_indices = sorted([int(selected\_layer.split('.')[-1][1:]) \PyDef{for} selected\_layer \PyDef{in} selected\_layers\_proj])}\\
\PyCode{indices = sum([list(range(dim\_head * head\_idx, dim\_head * (head\_idx + 1))) \PyDef{for} head\_idx \PyDef{in} head\_indices], [])}\\
\PyComment{Indices are split grouped by the dim\_head}\\
\PyComment{Projection-wise CA-LoRA (end)}\\

\PyCode{projection\_layer = modify\_to\_ca\_lora\_linear(projection\_layer, reduced\_layer='A' if is\_out else 'B')}\\
\PyComment{modify\_to\_ca\_lora\_linear: \Cref{algo:ca_lora_forward}}\\

\PyCode{\PyDef{if} is\_out:}

\Indp
\PyCode{projection\_layer.set\_lora\_layer(}\\
\PyCode{\quad LoRALinearLayer(}\\
\PyCode{\quad\quad in\_features=len(indices),}\\
\PyCode{\quad\quad out\_features=projection\_layer.out\_features,}\\
\PyCode{\quad\quad rank=rank),}\\
\PyCode{\quad indices)}
    
\Indm
\PyDef{else}:

\Indp
\PyCode{projection\_layer.set\_lora\_layer(}\\
\PyCode{\quad LoRALinearLayer(}\\
\PyCode{\quad\quad in\_features=projection\_layer.in\_features,}\\
\PyCode{\quad\quad out\_features=len(indices),}\\
\PyCode{\quad\quad rank=rank),}\\
\PyCode{\quad indices)}

\Indm
\Indm
\Indm

\PyDef{return} unet

\Indm

\end{algorithm*}




\end{document}